\documentclass[draft]{agujournal}
\draftfalse

\journalname{Geophysical Research Letters}

\usepackage{graphicx}

\graphicspath{{Figures/}}

\usepackage{amsmath}    
\usepackage{amssymb}
\usepackage{color, soul}

\newcommand{\coo}{${\rm {CO}_{2}}$}
\renewcommand{\vec}[1]{\boldsymbol{{#1}}} 

\newcommand{\cH}{\mathop{\mathcal{H}}}
\newcommand{\cG}{\mathop{\mathcal{G}}}

\newcommand{\cL}{\mathop{\mathcal{L}}}
\newcommand{\cS}{\mathop{\mathcal{S}}}

\newcommand{\be}{\begin{equation}}
\newcommand{\ee}{\end{equation}}

\begin{document}

%% ------------------------------------------------------------------------ %%
%  Title
% 
% (A title should be specific, informative, and brief. Use
% abbreviations only if they are defined in the abstract. Titles that
% start with general keywords then specific terms are optimized in
% searches)
%
%% ------------------------------------------------------------------------ %%

% Example: \title{This is a test title}

\title{Earth System Modeling 2.0: A Blueprint for Models That Learn From Observations and Targeted High-Resolution Simulations}

%% ------------------------------------------------------------------------ %%
%
%  AUTHORS AND AFFILIATIONS
%
%% ------------------------------------------------------------------------ %%

% Authors are individuals who have significantly contributed to the
% research and preparation of the article. Group authors are allowed, if
% each author in the group is separately identified in an appendix.)

% List authors by first name or initial followed by last name and
% separated by commas. Use \affil{} to number affiliations, and
% \thanks{} for author notes.  
% Additional author notes should be indicated with \thanks{} (for
% example, for current addresses). 

% Example: \authors{A. B. Author\affil{1}\thanks{Current address, Antartica}, B. C. Author\affil{2,3}, and D. E.
% Author\affil{3,4}\thanks{Also funded by Monsanto.}}

\authors{Tapio Schneider\affil{1,2}, Shiwei Lan\affil{1}, Andrew Stuart\affil{1} and Jo{\~a}o Teixeira\affil{2}}

% \affiliation{1}{First Affiliation}
% \affiliation{2}{Second Affiliation}
% \affiliation{3}{Third Affiliation}
% \affiliation{4}{Fourth Affiliation}

\affiliation{1}{California Institute of Technology, Pasadena, California 91125, USA}
\affiliation{2}{Jet Propulsion Laboratory, California Institute of Technology, Pasadena, California 91125, USA}
%(repeat as many times as is necessary)

%% Corresponding Author:
% Corresponding author mailing address and e-mail address:

% (include name and email addresses of the corresponding author.  More
% than one corresponding author is allowed in this LaTeX file and for
% publication; but only one corresponding author is allowed in our
% editorial system.)  

% Example: \correspondingauthor{First and Last Name}{email@address.edu}

\correspondingauthor{Tapio Schneider}{tapio@caltech.edu.\\
{\rm \copyright\ 2017. California Institute of Technology. Government sponsorship acknowledged.}}

%% Keypoints, final entry on title page.

% Example: 
% \begin{keypoints}
% \item	List up to three key points (at least one is required)
% \item	Key Points summarize the main points and conclusions of the article
% \item	Each must be 100 characters or less with no special characters or punctuation 
% \end{keypoints}

%  List up to three key points (at least one is required)
%  Key Points summarize the main points and conclusions of the article
%  Each must be 100 characters or less with no special characters or punctuation 

\begin{keypoints}
\item Earth system models (ESMs) and their parameterization schemes can be radically improved by data assimilation and machine learning
\item ESMs can learn from and integrate global observations from space and local high-resolution simulations
\item Ensemble Kalman inversion and Markov chain Monte Carlo methods show promise as learning algorithms for ESMs
\end{keypoints}

%% ------------------------------------------------------------------------ %%
%
%  ABSTRACT
%
% A good abstract will begin with a short description of the problem
% being addressed, briefly describe the new data or analyses, then
% briefly states the main conclusion(s) and how they are supported and
% uncertainties. 
%% ------------------------------------------------------------------------ %%

%% \begin{abstract} starts the second page 

\begin{abstract}
Climate projections continue to be marred by large uncertainties, which originate in processes that need to be parameterized, such as clouds, convection, and ecosystems. But rapid progress is now within reach. New computational tools and methods from data assimilation and machine learning  make it possible to integrate global observations and local high-resolution simulations in an Earth system model (ESM) that systematically learns from both and quantifies uncertainties. Here we propose a blueprint for such an ESM. We outline how parameterization schemes can learn from global observations and targeted high-resolution simulations, for example, of clouds and convection, through matching low-order statistics between ESMs, observations, and high-resolution simulations. We illustrate learning algorithms for ESMs with a simple dynamical system that shares characteristics of the climate system; and we discuss the opportunities the proposed framework presents and the challenges that remain to realize it.
\end{abstract}

%% ------------------------------------------------------------------------ %%
%
%  TEXT
%
%% ------------------------------------------------------------------------ %%

%%% Suggested section heads:
% \section{Introduction}
% 
% The main text should start with an introduction. Except for short
% manuscripts (such as comments and replies), the text should be divided
% into sections, each with its own heading. 

% Headings should be sentence fragments and do not begin with a
% lowercase letter or number. Examples of good headings are:

% \section{Materials and Methods}
% Here is text on Materials and Methods.
%
% \subsection{A descriptive heading about methods}
% More about Methods.
% 
% \section{Data} (Or section title might be a descriptive heading about data)
% 
% \section{Results} (Or section title might be a descriptive heading about the
% results)
% 
% \section{Conclusions}

\section{Introduction}

Climate models are built around models of the atmosphere, which are based on the laws of thermodynamics and on Newton's laws of motion for air as a fluid. Since they were first developed in the 1960s \citep{Smagorinsky63, Smagorinsky65, Manabe65, Mintz65a, Kasahara67a}, they have evolved from atmosphere-only models, via coupled atmosphere-ocean models with dynamic oceans, to Earth system models (ESMs) with dynamic cryospheres and biogeochemical cycles \citep{Bretherton12a, ipcc13}. Atmosphere and ocean models compute approximate numerical solutions to the laws of fluid dynamics and thermodynamics on a computational grid. For the atmosphere, the computational grid currently consists of $O(10^7)$ cells, spaced $O(10~\mathrm{km})$--$O(100~\mathrm{km})$ apart in the horizontal; for the oceans, the grid consists of $O(10^8)$ cells, spaced $O(10~\mathrm{km})$ apart in the horizontal. But scales smaller than the mesh size of a climate model cannot be resolved, yet are essential for its predictive capabilities. The unresolved scales are modeled by a variety of semi-empirical parameterization schemes, which represent the dynamics on subgrid-scales as parametric functions of the resolved dynamics on the computational grid \citep{Stensrud07a}. For example, the dynamical scales of stratocumulus clouds, the most common type of boundary layer clouds, are $O(10~\mathrm{m})$ and smaller, which will remain unresolvable on the computational grid of global atmosphere models for the foreseeable future \citep{Wood12a,Schneider17a}. Similarly, the submesoscale dynamics of oceans that may be important for biological processes near the surface have length scales of $O(100~\mathrm{m})$, which will also remain unresolvable for the foreseeable future \citep{Fox-Kemper14a}. Such smaller-scale dynamics in the atmosphere and oceans must be represented in climate models through parameterization schemes. Additionally, ESMs contain parameterization schemes for many processes for which the governing equations are not known or are only poorly known, for example, ecological or biogeochemical processes.

All of these parameterization schemes contain parameters that are uncertain, and the structure of the equations underlying them is uncertain itself. That is, there is parametric and structural uncertainty \citep{Draper95}. For example, entrainment and detrainment rates are parameters or parametric functions of state variables such as the vertical velocity of updrafts. They control the interaction of convective clouds with their environment and affect cloud properties and climate. But how they depend on state variables is uncertain, as is the structure of the closure equations in which they appear \citep[e.g.,][]{Stainforth05a,Holloway09a,Neelin09a,Romps10a,Nie12a,de-Rooy13a}. Or, as another example, the residence times of carbon in different reservoirs (e.g., soil, litter, plants) control how rapidly and where in the biosphere carbon accumulates. They affect the climate response of the biosphere. But they are likewise uncertain, differing by $O(1)$ factors among models \citep{Friedlingstein06a,Friedlingstein14a,Friend14a,Bloom16a}. Typically, parameterization schemes are developed and parameters in them are estimated independently of the  model into which they are eventually incorporated. They are tested with observations from field studies at a relatively small number of locations. For processes such as boundary-layer turbulence that are computable if sufficiently high resolution is available, parameterization schemes are increasingly also tested with data generated computationally in local process studies with high-resolution models \citep[e.g.,][]{Jakob03a,Jakob10a}. After the parameterization schemes are developed and incorporated in a climate model or ESM, modelers adjust (``tune'') parameters to satisfy large-scale  physical constraints, such as a closed energy balance at the top of the atmosphere (TOA), or selected observational constraints, such as reproduction of the 20th-century global-mean surface temperature record. This model tuning process currently relies on  knowledge and intuition of the modelers about plausible ranges of the tunable parameters and about the effect of parameter changes on the simulated climate of a model \citep{Randall97a,Mauritsen12a, Golaz13a,Hourdin13a, Flato13a, Hourdin17a}. But because of the nonlinear and interacting multiscale nature of the climate system, the simulated climate can depend sensitively and in unexpected ways on settings of tunable parameters \citep[e.g.,][]{Suzuki13a,Zhao16a}. It also remains unclear to what extent the resulting parameter choice is optimal, or how uncertain it is. Moreover, typically only a minute fraction of the available observations is used in the tuning process, usually only highly aggregated data such as global or large-scale mean values accumulated over periods of years or more. In part, this may be done to avoid overfitting, but more importantly, it is done because the tuning process usually involves parameter adjustments by hand, which each must be evaluated by a forward integration of the model. This makes the tuning process tedious and precludes adjustments of a larger set of parameters to fit more complex observational datasets or a wider range of high-resolution process simulations. It also precludes quantification of uncertainties \citep{Schirber13a,Hourdin17a}.

Climate models have improved over the past decades, leading, for example, to better simulations of El Ni{\~n}o, storm tracks, and tropical waves \citep{Guilyardi09a,Hung13a,Flato13a}. Weather prediction models, the higher-resolution siblings of the climate models' atmospheric component, have undergone a parallel evolution. Along with data assimilation techniques for the initialization of weather forecasts, this has led to great strides in the accuracy of weather forecasts \citep{Bauer15a}. But the accuracy of climate projections has not improved as much, and unacceptably large uncertainties remain.  For example, if one asks how high \coo\ concentrations can rise before Earth's surface will have warmed $2^\circ\mathrm{C}$ above pre-industrial temperatures---the warming target of the 2015 Paris Agreement, of which about $1^\circ\mathrm{C}$ remains because about $1^\circ\mathrm{C}$ has already been realized---the answers range from 480 to 600~ppm across current climate models \citep{Schneider17a}. A \coo\ concentration of 480~ppm will be reached in the late 2030s or early 2040s; 600~ppm may not be reached before 2060 even if \coo\ emissions continue to increase rapidly. Between these extremes lie vastly different optimal policy responses and socioeconomic costs of climate change \citep{Hope15a}.

These large and long-standing uncertainties in climate projections have their root in uncertainties in parameterization schemes. Parameterizations of clouds dominate the uncertainties in physical processes \citep{Cess89a,Cess90a,Stephens05,Bony06,Soden06a,Vial13a,Webb13b,Brient16b}. There are uncertainties both in the representation of the turbulent dynamics of clouds and in the representation of their microphysics, which control, for example, the distribution of droplet sizes in a cloud, the fraction of cloud condensate that precipitates out, and the phase partitioning of cloud condensate into liquid and ice \citep[e.g.,][]{Stainforth05a,Jiang12a,Suzuki13a,Golaz13a,Bodas-Salcedo14a,Zhao16a,Kay16a}. Additionally, there are numerous other parameterized processes that contribute to uncertainties in climate projections. For example, it is not precisely known what fraction of the \coo\ that is emitted by human activities will remain in the atmosphere, and so it is uncertain which emission pathways will lead to a given atmospheric \coo\ concentration target \citep{Knutti08b,Meinshausen09a,Friedlingstein15a}. Currently, only about half the emitted \coo\ accumulates in the atmosphere. The other half is taken up by oceans and on land. It is unclear in particular what fraction of the emitted \coo\ terrestrial ecosystems will take up in the future \citep{Friedlingstein06a,Canadell07a,Knorr09a,Le-Quere13a,Todd-Brown13a,Friedlingstein14a,Friend14a}. Reducing such uncertainties through the traditional approach to developing and improving parameterization schemes---attempting to develop one ``correct'' global parameterization scheme for each process in isolation, on the basis of observational or computational process studies that are usually focused on specific regions---has met only limited success \citep{Jakob03a,Jakob10a,Randall13b}.

Here we propose a new approach to improving parameterization schemes. The new approach invests considerable computational effort up front to exploit global observations and targeted high-resolution simulations through the use of data assimilation and machine learning within physical, biological, and chemical process models. We first outline in broad terms how we envision ESMs to learn from global observations and targeted high-resolution simulations (section~\ref{s:overview}). Then we discuss in more concrete terms the framework underlying such learning ESMs (section~\ref{s:learning}). We illustrate the approach by learning parameters in a relatively simple dynamical system that mimics characteristics of the atmosphere and oceans (section~\ref{s:lorenz96}). We conclude with an outlook of the opportunities the framework we outline presents and of the research program that needs to be pursued to realize it (section~\ref{s:discussion}).

\section{Learning from Observations and Targeted High-Resolution Simulations}\label{s:overview}

\subsection{Information Sources for Parameterization Schemes}

Parameterization schemes can learn from two sources of information:
\begin{enumerate}
\item \emph{Global observations}. We live in the golden age of Earth observations from space \citep{LEcuyer15a}. A suite of satellites flying in the formation known as the A-train has been streaming coordinated measurements of the composition of the atmosphere and of physical variables in the Earth system. We have nearly simultaneous measurements of variables such as temperature, humidity, and cloud and sea ice cover, with global coverage for more than a decade \citep{Stephens02a,Jiang12a,Simmons16a,Stephens17a}. Space-based measurements of biogeochemical tracers and processes, such as measurements of column-average \coo\ concentrations and of photosynthesis in terrestrial ecosystems, are also beginning to become available \citep[e.g.,][]{Crisp04a,Yokota09a,Frankenberg11a,Joiner11a,Frankenberg14a,Bloom16a,Eldering17a,Liu17a,Sun17a}, and so are more detailed observations of the cryosphere \citep[e.g.,][]{Shepherd12a,Gardner13a,Vaughan13a}. Parameterization schemes can learn from such space-based global data, which can be augmented and validated with more detailed local observations from the ground and from field studies.
\item \emph{Local high-resolution simulations}. Some processes parameterized in ESMs are in principle computable, only the globally achievable resolution precludes their explicit computation. For example, the turbulent dynamics (though currently not the microphysics) of clouds can be computed with high fidelity in limited domains in large-eddy simulations (LES) with grid spacings of $O(10~\mathrm{m})$ \citep{Siebesma03,Stevens05a,Khairoutdinov09a,Matheou14a,Schalkwijk15a,Pressel15a,Pressel17a}. Increased computational performance has made LES domain widths of $O(10~\mathrm{km})$--$O(100~\mathrm{km})$ feasible in recent years, while the horizontal mesh size in atmosphere models has shrunk, to the point that the two scales have converged. Thus, while global LES that reliably resolve low clouds such as cumulus or stratocumulus will not be feasible for decades, it is possible to nest LES in selected grid columns of atmosphere models and conduct high-fidelity local simulations of cloud dynamics in them \citep{Schneider17a}. Local high-resolution simulations of ocean mesoscale turbulence or sea ice dynamics can be conducted similarly. Parameterization schemes can learn from such nested high-resolution simulations.
\end{enumerate}
Of course, both observations and high-resolution simulations have been exploited in the development of parameterization schemes for some time. For example, data assimilation techniques have been used to estimate parameters in parameterization schemes from observations. Parameters especially in cloud, convection, and precipitation parameterizations have been estimated by minimizing errors in short-term weather forecasts over timescales of hours or days \citep[e.g.,][]{Emanuel99a,Grell02a,Aksoy06a,Schirber13a,Ruiz13a,Ruiz15a}, or by minimizing deviations between simulated and observed longer-term aggregates of climate statistics, such as global-mean TOA radiative fluxes accumulated over seasons or years \citep[e.g.,][]{Jackson08a,Jarvinen10a,Neelin10a,Solonen12a,Tett13a}. High-resolution simulations have been used to provide detailed dynamical information such as vertical velocity and turbulence kinetic energy profiles in convective clouds, which are not easily available from observations. They have often been employed to augment observations from local field studies, and parameterization schemes have been fit to and evaluated with the observations and the high-resolution simulations used in tandem \citep[e.g.,][]{Liu01a,Siebesma03,Stevens05a,Siebesma07,Hohenegger11a,de-Rooy13a,Romps16a}. High-resolution deep convection-resolving simulations with $O(1~\mathrm{km})$ horizontal grid spacing and, most recently, LES with $O(100~\mathrm{m})$ horizontal grid spacing have also been nested in small, usually two-dimensional subdomains of atmospheric grid columns, as a parameterization surrogate that explicitly resolves some aspects of cloud dynamics \citep[e.g.,][]{Grabowski99a,Grabowski01a,Khairoutdinov01a,Randall03a,Khairoutdinov05a,Randall13b,Grabowski16a,Parishani17a}. Such multiscale modeling approaches, often called superparameterization, have led to markedly improved simulations, for example, of the Asian monsoon, of tropical surface temperatures, and of precipitation and its diurnal cycle, albeit at great computational expense \citep[e.g.,][]{Benedict09a,Pritchard09a,Pritchard09b,Stan10a,DeMott13a}. However, multiscale modeling relies on a scale separation between the global-model mesh size and the domain size of the nested high-resolution simulation \citep{E07a}. Multiscale modeling is computationally advantageous relative to global high-resolution simulations as long as it suffices for the nested high-resolution simulation to subsample only a small fraction of the footprint of a global-model grid column, and to extrapolate the information so obtained to the entire footprint on the basis of statistical homogeneity assumptions. As the mesh size of global atmosphere models shrinks to horizontal scales of kilometers---resolutions that are already feasible in short integrations or limited areas and that will become routine in the next decade \citep{Palmer14c,Ban15a,Ohno16a,Schneider17a}---the scale separation to the minimum necessary domain size of nested high-resolution simulations will disappear, and with it the computational advantage of multiscale modeling.

What we propose here combines elements of these existing approaches in a novel way. At its core are still parameterization schemes that are based on physical, biological, or chemical process models, whose mathematical structure is developed on the basis of theory, local observations, and, where possible, high-resolution simulations. But we propose that these parameterization schemes, when they are embedded in ESMs, learn directly from observations and high-resolution simulations that both sample the globe. High-resolution simulations are employed in a targeted way---akin to targeted or adaptive observations in weather forecasting \citep{Palmer98a,Lorenz98a,Bishop01a}---to reduce uncertainties where observations are insufficient to obtain tight parameter estimates. Instead of incorporating high-resolution simulations globally in a small fraction of the footprint of each grid column like in multiscale modeling approaches, the ESM we envision deploys them locally, in entire grid columns, albeit only in a small subset of them. High-resolution simulations can be targeted to grid columns selected based on measures of uncertainty about model parameters. If the nested high-resolution simulations feed back onto the ESM, this corresponds to a locally extreme mesh refinement; however, two-way nesting may not always be necessary \citep[e.g.,][]{Moeng07a,Zhu10a}. The model learns parameters from observations and from nested high-resolution simulations in a computationally intensive learning phase, after which it can be used in a computationally more efficient manner, like models in use today. Nonetheless, even in simulations of climates beyond what has been observed, bursts of targeted high-resolution simulations can continue to be deployed to refine parameters and estimate their uncertainties. 

\subsection{Computable and Non-computable Parameters}

Learning from  high-resolution simulations and observations is aimed at determining two different kinds of parameters in parameterization schemes: \emph{computable} and \emph{non-computable} parameters. (Since parameters and parametric functions of state variables play essentially the same role in our discussion, we simply use the term parameter, with the understanding that this can include parametric functions and even nonparametric functions.) Computable parameters are those that can in principle be inferred from high-resolution simulations alone. They include parameters in radiative transfer schemes, which can be inferred from detailed line-by-line calculations; dynamical parameters in cloud turbulence parameterizations, such as entrainment rates, which can be inferred from LES; or parameters in ocean mixing parameterizations, which can be inferred from high-resolution simulations. Non-computable parameters are parameters that, currently, cannot be inferred from high-resolution simulations, either because computational limitations make it necessary for them to also appear in parameterization schemes in high-resolution simulations, or because the microscopic equations governing the processes in question are unknown. They include parameters in cloud microphysics parameterizations, which are still necessary to include in LES, and many parameters characterizing ecological and biogeochemical processes, whose governing equations are unknown. Cloud microphysics parameters will increasingly become computable through direct numerical simulation \citep{Devenish12a,Grabowski13a}, but ecological and biogeochemical parameters will remain non-computable for the foreseeable future. Both computable and non-computable parameters can, in principle, be learned from observations; the only restrictions to their identifiability come  from the well-posedness of the learning problem and its computational tractability. But only computable parameters can be learned from targeted high-resolution simulations. To be able to learn computable parameters, it is essential to represent non-computable aspects of a parameterization scheme consistently in the high-resolution simulation and in the parameterization scheme that is to learn from the high-resolution simulation. For example, radiative transfer and microphysical processes need to be represented consistently in a high-resolution LES and in a parameterization scheme if the parameterization scheme is to learn computable dynamical parameters such as entrainment rates from the LES. 

This approach presents challenges for parameter learning, since it implies the need to use observational data and high-resolution simulations in tandem to improve model parameterizations. But it also presents an opportunity: in doing so, the reliability and predictive power of ESMs can be improved, and uncertainties in parameters and predictions can be quantified.

\subsection{Objectives: Bias Reduction and Exploitation of Emergent Constraints}

Computational tractability is paramount for the success of any parameter learning algorithm for ESMs \citep[e.g.,][]{Annan07a,Jackson08a,Neelin10a,Solonen12a}. The central issue is the number of times the objective function needs to be evaluated, and hence an ESM needs to be run, in the process of parameter learning. Standard parameter estimation and inverse problem approaches may require $O(10^5)$ function or  derivative evaluations to learn $O(100)$ parameters, especially if uncertainty in the estimates is also required \citep{Cotter13a}. This many forward integrations and/or derivative evaluations of ESMs are not feasible if each involves accumulation of longer-term climate statistics. Fast parameterized processes in climate models often exhibit errors within a few hours or days of integration that are similar to errors in the mean state of the model \citep{Phillips04a,Rodwell07a,Xie12a,Ma13a,Klocke14a}. This has given rise to hopes that it may suffice to evaluate objective functions by weather hindcasts over timescales of only hours, making many evaluations of an objective function feasible \citep{Aksoy06a,Ruiz13a,Wan14a}. But experience has shown that such short-term optimization may not always lead to the desired improvements in climate simulations \citep{Schirber13a}. Additionally, slower parameterized processes, for example, involving biogeochemical cycles or the cryosphere, require longer integration times to accumulate statistics entering any meaningful objective function. Therefore, we focus on objective functions involving climate statistics accumulated over windows that we anticipate to be wide compared with the $O(10~\mathrm{days})$ timescale over which the atmosphere forgets its initial condition. Then the accumulated statistics do not depend sensitively on atmospheric initial conditions. This reduces the onus of correctly assimilating atmospheric initial conditions in parameter learning, which would be required if one were to match simulated and observed trajectories, as in approaches that assimilate model parameters jointly with the state of the system by augmenting state vectors with parameters \citep[e.g.,][]{Dee05a,Aksoy06a,Anderson09a}. The minimum window over which climate statistics will need to be accumulated will vary from processes to process, generally being longer for slower processes (e.g., the cryosphere) than faster processes (e.g., the atmosphere). For slower processes whose initial condition is not forgotten over the accumulation window, it will remain necessary to correctly assimilate initial conditions. 

The objective functions to be minimized in the learning phase can be chosen to directly minimize biases in climate simulations, for example, precipitation biases such as the longstanding double-ITCZ bias in the tropics  \citep{Lin07a,Li14a,Zhang15a,Adam16b,Adam17a}, or cloud cover biases such as the ``too few--too bright'' bias in the subtropics \citep{Webb01a,Zhang05b,Karlsson08a,Nam12a}. Because the sensitivity with which an ESM responds to increases in greenhouse gas concentrations correlates with the spatial structure of some of these biases in the models \citep[e.g.,][]{Tian15a,Siler17a}, minimizing regional biases will likely reduce uncertainties in climate projections, in addition to leading to more reliable simulations of the present climate. To minimize biases, the objective function needs to include mean-field terms penalizing mismatch between spatially and at least seasonally resolved simulated and observed mean fields, for example, of precipitation, ecosystem primary productivity, and TOA radiative energy fluxes. 

Additionally, there is a growing literature on ``emergent constraints,'' which typically are fluctuation-dissipation relationships that relate measurable fluctuations in the present climate to the response of the climate system to perturbations \citep{Hall06a,Collins12a,Klein15a}. For example, how strongly tropical low-cloud cover covaries with surface temperature from year to year or even seasonally in the present climate correlates in climate models with the amplitude of the cloud response to global warming  \citep{Qu14a,Qu15a,Brient16b}. Therefore, the observable low-cloud cover covariation with surface temperature in the present climate can be used to constrain the cloud response to global warming. Or, as another example, how strongly atmospheric \coo\ concentrations covary with surface temperature in the present climate correlates in climate models with the amplitude of the terrestrial ecosystem response to global warming (e.g., the balance between \coo\ fertilization of plants and enhanced soil and plant respiration under warming) \citep{Cox13a,Wenzel14a}. Therefore, the observable \coo\ concentration covariation with surface temperature can be used to constrain the terrestrial ecosystem response to global warming. Such emergent constraints are usually used post facto, in the evaluation of ESMs. They lead to inferences about the likelihood of a model given the measured natural variations, and they therefore can be used to assess how likely it is that its climate change projections are correct \citep[e.g.,][]{Brient16b}. But emergent constraints usually are not used directly to improve models. In what we propose, they are used directly to learn parameters in ESMs and to reduce uncertainties in the climate response. To do so, covariance terms (e.g., between surface temperature and cloud cover or TOA radiative fluxes, or between surface temperature and \coo\ concentrations) need to be included in the objective function. 

The choice of objective functions to be employed is key to the success of what we propose. The use of time-averaged statistics such as mean-field and covariance terms will make the objective functions smoother and hence reduce the computational cost of minimization, compared with minimizing objective functions that directly penalize mismatch between simulated and observed trajectories of the Earth system. From the point of view of statistical theory, the objective functions should contain the sufficient statistics for the parameters of interest, but what these are is not usually known a priori. In practice, the choice of objective functions will be guided by expertise specific to the relevant subdomains of Earth system science, as well as computational cost. Given that current ESM components such as clouds and the carbon cycle exhibit large seasonal biases \citep[e.g.,][]{Keppel-Aleks12a,Karlsson13a,Lin14b}, and their response to long-term warming in some respects resembles their response to seasonal variations \citep[e.g.,][]{Brient16b, Wenzel16a}, accumulating seasonal statistics in the objective functions suggests itself as a starting point.

\section{Machine Learning Framework for Earth System Models}\label{s:learning}

\subsection{Models and Data}

To outline how we envision parameterization schemes in ESMs to learn from diverse data, we first set up notation. Let $\vec{\theta}=(\vec{\theta}_c, \vec{\theta}_n)$ denote the vector of model parameters to be learned, consisting of computable parameters $\vec{\theta}_c$ that can be learned from high-resolution simulations, and non-computable parameters $\vec{\theta}_n$ that can only be learned from observations (for example, because high-resolution simulations themselves depend on $\vec{\theta}_n$). The parameters $\vec{\theta}$ appear in parameterization schemes in a model, which may be viewed as a map $\cG$, parameterized by time $t$, that takes the parameters $\vec{\theta}$ to the state variables $\vec{x}$, 
\begin{equation}
\vec{x}(t) = \cG(\vec{\theta}, t).
\end{equation}
The state variables $\vec{x}$ can include temperatures, humidity variables, and cloud, cryosphere, and biogeochemical variables, and the map $\cG$ may depend on initial conditions and time-evolving boundary or forcing conditions. The map $\cG$ typically represents a global ESM. The state variables $\vec{x}$ are linked to observables $\vec{y}$ through a map $\cH$ representing an observing system, so that 
\begin{equation}
\vec{y}(t)=\cH\bigl(\vec{x}(t)\bigr).
\end{equation}
The observables $\vec{y}$ might represent surface temperatures, \coo\ concentrations, or spectral radiances emanating from the TOA. The map $\cH$ in practice will be realized through an observing system simulator, which simulates how observables $\vec{y}$ are impacted by a multitude of state variables $\vec{x}$. The actual observations (e.g., space-based measurements) are denoted by $\vec{\tilde y}$, so $\vec{y}(t) - \vec{\tilde y}(t)$ is the mismatch between simulations and observations. Since $\vec{y}$ is parameterized by $\vec{\theta}$, while $\vec{\tilde y}$ is independent of $\vec{\theta}$, mismatches between $\vec{y}$ and  $\vec{\tilde y}$ can be used to learn about $\vec{\theta}$. 

Local high-resolution simulations nested in a grid column of an ESM may be viewed as a time-dependent map $\cL$ from the state variables $\vec{x}$ of the ESM to simulated state variables $\vec{\tilde z}$,
\begin{equation}
\vec{\tilde z}(t) = \cL(\vec{\theta}_n,t; \vec{x}).
\end{equation}
The map $\cL$ is parameterized by non-computable parameters $\vec{\theta}_n$ and time $t$, and it can involve the time-history of the state variables $\vec{x}$ up to time $t$. The vector $\vec{\tilde z}$ contains statistics of high-resolution variables whose counterparts in the ESM are computed by parameterization schemes, such as the mean cloud cover or liquid water content in a grid box. The corresponding variables $\vec{z}$ in the ESM are obtained by a time-dependent map $\cS$ that takes state variables $\vec{x}$ and parameters $\vec{\theta}$ to $\vec{z}$,
\begin{equation}
\vec{z}(t) = \cS(\vec{\theta},t; \vec{x}).
\end{equation}
The map $\cS$ typically represents a single grid column of the ESM with its parameterization schemes, taking as input $\vec{x}$ from the ESM. It is structurally similar to $\cL$. Crucially, however, $\cS$ generally depends on all parameters $\vec{\theta}=(\vec{\theta}_c, \vec{\theta}_n)$, while $\cL$ only depends on non-computable parameters $\vec{\theta}_n$. Thus, the mismatch $\vec{z}(t) - \vec{\tilde z}(t)$ can be used to learn about the computable parameters $\vec{\theta}_c$.

The same framework also covers other ways of learning about parameterizations schemes from data. For example, the map $\cG$ may represent a single grid column of an ESM, driven by time-evolving boundary conditions from reanalysis data at selected sites. Observations at the sites can then be used to learn about the parameterization schemes in the column \citep{Neggers12a}. Or, similarly, the map $\cG$ may represent a local high-resolution simulation driven by reanalysis data, with parameterization schemes, e.g., for cloud microphysics, about which one wants to learn from observations. 

\subsection{Objective Functions}

Objective functions are defined through mismatch between the simulated data $\vec{y}$ and observations $\vec{\tilde y}$, on the one hand, and simulated data $\vec{z}$ and high-resolution simulations $\vec{\tilde z}$, on the other hand. We define mismatches using time-averaged statistics because they do not suffer from sensitivity to atmospheric initial conditions; indeed, matching trajectories directly requires assimilating atmospheric initial conditions, which would make it difficult to disentangle mismatches due to errors in climatically unimportant atmospheric initial conditions from those due to parameterization errors. However, the time averages can still depend on initial conditions for slowly evolving components of the Earth system, such as ocean circulations or ice sheets.

We denote the time average of a function $\phi(t)$ over the time interval $[t_0,t_0+T]$ by
\begin{equation}\label{e:Tavg}
\langle \phi \rangle_T = \frac{1}{T} \int_{t_0}^{t_0+T} \phi(t) \, dt.
\end{equation}
The observational objective function can then be written in the generic form
\begin{equation}\label{e:obj_o}
J_o(\vec{\theta})=\frac12\| \langle \vec{f}(\vec{y})  \rangle_T - \langle \vec{f}(\vec{\tilde y})
\rangle_T \|_{\Sigma_y}^2
\end{equation}
with the 2-norm
\begin{equation}
\|\cdot\|_{\Sigma_y}=\|\Sigma_y^{-\frac12}\cdot\|
\end{equation}
normalized by error standard deviations and covariance information captured in $\Sigma_y$. The function
$\vec{f}$ of the observables typically involves first- and second-order quantities, for example,
\begin{equation}
\vec{f}(\vec{y}) = \left( 
\begin{array}{c} 
\vec{y}\\
y_i' y_j'
\end{array}
\right),
\label{e:of}
\end{equation}
where, for any observable $\phi$, $\phi'(t) = \phi(t) - \langle \phi \rangle_T$ denotes the fluctuation of $\phi$ about its mean $\langle \phi \rangle_T$. With $\vec{f}$ given by \eqref{e:of}, the objective function penalizes mismatch between the vectors of mean values $\langle \vec{y} \rangle_T$ and $\langle \vec{\tilde y}\rangle_T$ and between the covariance components $\langle y_i' y_j' \rangle_T$ and  $\langle \tilde y_i' \tilde y_j' \rangle_T$ for some indices $i$ and $j$. The least-squares form of the objective function \eqref{e:obj_o} follows from assuming an error model
\begin{equation}
\vec{f}(\vec{y})=\vec{f}(\vec{\tilde y})+\vec{\eta},
\end{equation}
with the matrix $\Sigma_y$ encoding an assumed covariance structure of the noise vector $\vec{\eta}$. The relevant components of $\Sigma_y$ may be chosen very small for quantities that are used as constraints on the ESM (e.g., the requirement of a closed global energy balance at TOA).

For the mismatch to high-resolution simulations, we accumulate statistics over an ensemble of high-resolution simulations in different grid columns of the ESM and at different times, possibly, but not necessarily, also accumulating in time. We denote the corresponding ensemble and time average by $\langle \phi \rangle_E$, and define an objective function analogously to that for the observations through
\begin{equation}\label{e:obj_hr}
J_s(\vec{\theta}_c)=\frac12\| \langle \vec{g}(\vec{z})  \rangle_E - \langle \vec{g}(\vec{\tilde z})
\rangle_E \|_{\Sigma_z}^2.
\end{equation}
Like the function $\vec{f}$ above, the function $\vec{g}$  typically involves first- and second-order quantities, and the least-squares form of the objective functions follows from the assumed covariance structure $\Sigma_z$ of the noise. 

\subsection{Learning Algorithms}

Learning algorithms attempt to choose parameters $\vec{\theta}$ that minimize $J_o$ and $J_s$. However, minimization of $J_o$ and $J_s$ does not always determine the parameters uniquely, for example, if there are strongly correlated parameters or if the number of parameters to be learned exceeds the number of available observational degrees of freedom. In such cases, regularization is necessary to choose a good solution for the parameters among the multitude of possible solutions. This may be achieved in various ways: by adding to the least-squares objective functions \eqref{e:obj_o} and \eqref{e:obj_hr} regularizing penalty terms that incorporate prior knowledge about the parameters \citep{Engl96}, by Bayesian probabilistic regularization \citep{Kaipio05a}, or by restriction of the parameters to a subset, as in ensemble Kalman inversion \citep{Iglesias13a}. 

All of these regularization approaches may be useful in ESMs. They involve different trade-offs between computational expense and the amount of information about the parameters they provide. 
\begin{itemize}
\item Classical regularized least squares leads to an optimization problem that is typically tackled by gradient descent or Gauss-Newton methods, in which derivatives of the parameter-to-data map are employed \citep{Nocedal06a}. Such methods usually require $O(10^2)$ integrations of the forward model or evaluations of its derivatives with respect to parameters. 
\item Bayesian inversions usually employ Markov chain Monte Carlo (MCMC) methods \citep{Brooks11a} and variants such as sequential Monte Carlo \citep{Del-Moral06a} to approximate the posterior probability density function (PDF) of parameters, given data and a prior PDF. A PDF of parameters provides much more information than a point estimate, and consequently MCMC methods typically require many more forward model integrations, sometimes on the order of $O(10^5)$. The computational demands can be decreased by an order of magnitude by judicious use of derivative information where available (see \citet{Beskos17a} and references therein) or by improved sampling strategies \citep[e.g.,][]{Jackson08a,Jarvinen10a,Solonen12a}. Nonetheless, the cost remains orders of magnitude higher than for optimization techniques. 
\item Ensemble Kalman methods are easily parallelizable, derivative-free alternatives to the classical optimization and Bayesian approaches \citep{Houtekamer16a}. Although theory for them is less well developed, empirical evidence demonstrates behavior similar to derivative-based algorithms in complex inversion problems, with a comparable number of forward model integrations \citep{Iglesias16a}. Ensemble methods for joint state and parameter estimation have recently been systematically developed \citep{Bocquet13a,Bocquet14a,Carrassi17a}, and they are emerging as a promising way to solve inverse problems and to obtain qualitative estimates of uncertainty. However, numerical experiments have indicated that such uncertainty information is qualitative at best: the Kalman methods invoke Gaussian assumptions, which may not be justified, and even if the Gaussian approximation holds, the ensemble sizes needed for uncertainty quantification may not be practical \citep{Law12a,Iglesias13a}. 
\end{itemize}

An important consideration is how to blend the information about parameters contained in the high-resolution simulations and in the observations. One approach is as follows, although others may turn out to be preferable. Minimizing the high-resolution objective function $J_s$ in principle gives the computable parameters $\vec{\theta}_c$ as an implicit function of the non-computable parameters $\vec{\theta}_n$. This implicit function may then be used as prior information to minimize  the observational objective function $J_o$ over $\vec{\theta}.$ Bayesian MCMC approaches may be feasible for fitting $J_s$, since the single column model $\cS$ is relatively cheap to evaluate, and the ensemble of high-resolution simulations $\cL$ needed may not be large. Although Bayesian approaches may not be feasible for fitting $J_o$, for which accumulation of statistics of the model $\cG$ is required, this hierarchical approach does have the potential to incorporate detailed uncertainty estimates coming from the high-resolution simulations.

The choice of normalization (i.e., $\Sigma_y$ and $\Sigma_z$) in the objective functions plays a significant role in parameter learning, and learning about it has been demonstrated to have considerable impact on data assimilation for weather forecasts \citep{Dee95a,Stewart14a}. We will not discuss this issue in any detail, but note it may be addressed by the use of hierarchical Bayesian methodology and ensemble Kalman analogues. Nor will we dwell on the important issue of structural uncertainty---model error--- other than to note that this can, in principle, be addressed through the inverse problem approach advocated here: additional unknown parameters, placed judiciously within the model to account for model error, can be learned from data \citep{Kennedy01a,Dee05a}. The choice of normalization is especially important in this context as it relates to disentangling learning about model error from learning about the other parameters of interest.

Learning algorithms for ESMs can be developed further in several ways:
\begin{itemize}
\item Minimization of the objective functions $J_o$ and $J_s$ may be performed by online filtering algorithms, akin to those used in the initialization of weather forecasts, which sequentially update parameters as information becomes available \citep{Law15a}. This can reduce the number of forward model integrations required for parameter estimation, and it can allow parameterization schemes to learn adaptively from high-resolution simulations during the course of a global simulation. 
\item Where to employ targeted high-resolution simulations can be chosen to optimize aspects of the learning process. The simplest approach would be to deploy them randomly, for example, by selecting regions with a probability proportional to their climatological cloud fraction for high-resolution simulations of clouds. More efficient would be techniques of optimal experimental design (see \citet{Alexanderian16a}
and references therein), within online filtering algorithms. With such techniques, high-resolution simulations could be generated to order, to update aspects of parameterization schemes that have the most influence on the global system with which they interact. 
\end{itemize}
Progress along these lines will require innovation. For example, filtering algorithms need to be adapted to deal with strong serial correlations such as those that arise when averages $\langle \phi \rangle_{T_i}$ are accumulated over increasing spans $T_i < T_{i+1}$ and parameters are updated from one average $\langle \phi \rangle_{T_i}$ to a longer average $\langle \phi \rangle_{T_{i+1}}$. And optimal experimental design techniques require the development of cheap computational methods to evaluate sensitivities of the ESM to individual aspects of parameterization schemes.
  
\section{Illustration With Dynamical System}\label{s:lorenz96}

We envision ESMs eventually to learn parameters online, with targeted high-resolution simulations triggering parameter updates on the fly. Here we want to illustrate in off-line mode some of the opportunities and challenges of learning parameters in a relatively simple dynamical system. We use the Lorenz-96 model \citep{Lorenz96a}, which has nonlinearities resembling the advective nonlinearities of fluid dynamics and a multiscale coupling of slow and fast variables similar to what is seen in ESMs. The model has been used extensively in the development and testing of data assimilation methods \citep[e.g.,][]{Lorenz98a, Anderson01a, Ott04a}. 

\subsection{Lorenz-96 Model}

The Lorenz-96 model consists of $K$ slow variables $X_k$ ($k=1, \dots, K$), each of which is coupled to $J$ fast variables $Y_{j,k}$ ($j=1,\dots, J$) \citep{Lorenz96a}:
\begin{align}
\frac{dX_k}{dt}     &= - X_{k-1}(X_{k-2} - X_{k+1}) - X_k + F - hc \bar{Y}_k, \label{e:l96_slow}\\
\frac{1}{c} \, \frac{dY_{j,k}}{dt} &= -b Y_{j+1,k}(Y_{j+2,k} - Y_{j-1, k}) - Y_{j,k} + \frac{h}{J} X_k. \label{e:l96_fast}
\end{align}
The overbar denotes the mean value over $j$,
\begin{equation}
\bar{Y}_k = \frac{1}{J} \sum_{j=1}^J Y_{j,k}.
\end{equation}
Both the slow and fast variables are taken to be periodic in $k$ and $j$, forming a cyclic chain with $X_{k+K} = X_k$, $Y_{j, k+K}=Y_{j, k}$, and $Y_{j+J,k} = Y_{j,k+1}$.  The slow variables $X$ may be viewed as resolved-scale variables and the fast variables $Y$ as unresolved variables in an ESM. Each of the $K$ slow variables $X_k$ may represent a property such as surface air temperature in a cyclic chain of grid cells spanning a latitude circle. Each slow variable $X_k$ affects the $J$ fast variables $Y_{j,k}$ in the grid cell, which might represent cloud-scale variables such as liquid water path in each of $J$ cumulus clouds. In turn, the mean value of the fast variables over the cell, $\bar{Y}_k$, feeds back onto the slow variables $X_k$. The strength of the coupling between fast and slow variables is controlled by the parameter $h$, which represents an interaction coefficient, for example, an entrainment rate that couples cloud-scale variables to their large-scale environment. Time is nondimensionalized by the linear-damping timescale of the slow variables, which we nominally take to be 1~day, a typical thermal relaxation time of surface temperatures \citep{Swanson97b}. The parameter $c$ controls how rapidly the fast variables are damped relative to the slow; it may be interpreted as a microphysical parameter controlling relaxation of cloud variables, such as a precipitation efficiency.  The parameter $F$ controls the strength of the external large-scale forcing, and $b$ the amplitude of the nonlinear interactions among the fast variables. Following \citet{Lorenz96a}, albeit relabeling parameters, we choose $K=36$, $J=10$, $h=1$, and $F=c=b=10$, which ensures chaotic dynamics of the system. 

The quadratic nonlinearities in this dynamical system resemble advective nonlinearities, e.g., in the sense that they conserve the quadratic invariants (``energies'')  $\sum_{k} X_k^2$ and $\sum_{j} Y_{j, k}^2$ \citep{Lorenz98a}.  The interaction between the slow and fast variables conserves the ``total energy'' $\sum_k \left(X_k^2 + \sum_j Y_{j, k}^2\right)$. Energies are damped by the linear terms; they are prevented from decaying to zero by the external forcing $F$. Eventually, the system approaches a statistically steady state in which driving by the external forcing $F$ balances the linear damping. 

Let $\langle \cdot \rangle_\infty$ denote a long-term time mean in the statistically steady state, and note that all slow variables $X_k$ are statistically identical, as are all fast variables $Y_{j,k}$, so we can use the generic symbols $X$ and $Y$ in statistics of the variables. Multiplication of \eqref{e:l96_slow} by $X_k$, using that all variables $X_k$ are statistically identical, and averaging shows that, in the statistically steady state, second moments of the slow variables satisfy
\begin{equation}\label{e:moments1}
\langle X^2 \rangle_\infty = F \langle X \rangle_\infty - hc \langle X \bar Y \rangle_\infty.
\end{equation}
Similarly, second moments of the fast variables satisfy 
\begin{equation}\label{e:moments2}
\langle \overline{Y^2} \rangle_\infty = \frac{h}{J} \langle X \bar Y \rangle_\infty,
\end{equation}
where the overbar again denotes a mean value over the fast-variable index $j$. That is, the interaction coefficient $h$ can be determined from estimates of the one-point statistics $\langle \overline{Y^2} \rangle_\infty$ and $\langle X \bar Y \rangle_\infty$. Its inverse is proportional to the regression coefficient of the fast variables onto the slow: $h^{-1} \propto \langle X \bar Y \rangle_\infty/\langle \overline{Y^2} \rangle_\infty$. So the regression of the fast variables onto the slow can be viewed as providing an ``emergent  constraint'' on the system, insofar as the interaction coefficient $h$ affects the response of the system to perturbations (e.g., in $F$). Estimates of $\langle X^2 \rangle_\infty$ and $\langle X \rangle_\infty$ provide an additional constraint \eqref{e:moments1} on the parameters $F$ and $c$. Taking mean values of the dynamical equations \eqref{e:l96_slow} and \eqref{e:l96_fast} would provide further constraints on these parameters, as well as on $b$, in terms of two-point statistics involving shifts in $k$ and $j$, e.g., covariances of $X_k$ and $X_{k-1}$. 

In what follows, we demonstrate the performance of learning algorithms in a perfect-model setting, first focusing on one-point statistics to show how to learn about parameters in the full dynamical system from them. Subsequently, we use two-point statistics to learn about parameters in a single ``grid column'' of fast variables only.

\subsection{Parameter Learning in Perfect-Model Setting}

We generate data from the dynamical system \eqref{e:l96_slow} and \eqref{e:l96_fast} with the parameters $\vec{\theta} = (F, h, c, b)$ set to $\vec{\tilde\theta} = (10, 1, 10, 10)$. The role of ``observations'' $\vec{\tilde y} = (\tilde X, \tilde Y)$ in the perfect-model setting is played by data $\tilde X$ and $\tilde Y$ generated by the dynamical system with parameters $\vec{\theta}$ set to their ``true'' values $\vec{\tilde\theta}$. That is, the dynamical system \eqref{e:l96_slow} and \eqref{e:l96_fast} with parameters $\vec{\theta}$ stands for the global model $\cG$, the observing system map $\cH$ is the identity, and the data $\tilde X$ and $\tilde Y$ generated by the dynamical system with parameters $\vec{\tilde \theta}$ is a surrogate for observations. The parameters $\vec{\theta}$ of the dynamical system are then learned by matching statistics $\langle \phi \rangle_T$ accumulated over $T = 100~\mathrm{days}$ (with 1~day denoting the unit time of the system),  using discrete sums in place of the time integral in the average \eqref{e:Tavg} and minimizing the ``observational'' objective function 
\begin{equation}\label{e:obj_o1}
J_o(\vec{\theta})=\frac12\| \langle \vec{f}(X, Y)  \rangle_T - \langle \vec{f}(\tilde X, \tilde Y)
\rangle_\infty \|_{\Sigma}^2.
\end{equation}
The moment function to be matched,
\begin{equation}
\vec{f}(X, Y) = \left( 
\begin{array}{c} 
X\\
\bar Y\\
X^2\\
X\bar Y\\
\overline{Y^2}
\end{array}
\right),
\end{equation}
has an entry for each of the $K=36$ indices $k$, giving a vector of length $5K =180$. The noise covariance matrix $\Sigma$ is chosen to be diagonal, with entries that are proportional to the sample variances of the moments contained in the vector $\vec{f}$,
\begin{equation}\label{e:noise}
\mathop{\mathrm{diag}} \Sigma = r^2 \bigl[\mathop{\mathrm{var}}(X), \mathop{\mathrm{var}}({\bar Y}), \mathop{\mathrm{var}}({X^2}), \mathop{\mathrm{var}}({X\bar Y}), \mathop{\mathrm{var}}({\overline{Y^2}})\bigr].
\end{equation}
Here $\mathop{\mathrm{var}}(\phi)$ denotes the variance of $\phi$, and $r$ is an empirical parameter indicating the noise level. The variances $\mathop{\mathrm{var}}(\phi)$ and the ``true moments'' $\langle \vec{f}(\tilde X, \tilde Y)\rangle_\infty$ are estimated from a long (46,416~days) control simulation of the dynamical system with the true parameters $\vec{\tilde\theta}$.

As an illustrative example, we use normal priors for $(\theta_1, \theta_2, \theta_4) = (F, h, b)$, with mean values $(\mu_1, \mu_2, \mu_4) = (10, 0, 5)$ and variances $(\sigma^2_1, \sigma^2_2, \sigma^2_4) = (10, 1, 10)$. Enforcing positivity of $c$, we use a log-normal prior for $\theta_3 = c$, with a mean value $\mu_3 = 2$ and variance $\sigma_3^2 = 0.1$ for $\log c$ (i.e., a mean value of $7.4$ for $c$). We take the parameters a priori to be uncorrelated, so that the prior covariance matrix is diagonal. 

\begin{figure*}[htb] %  figure placement: here, top, bottom, or page
   \centering
   \includegraphics[width=160mm]{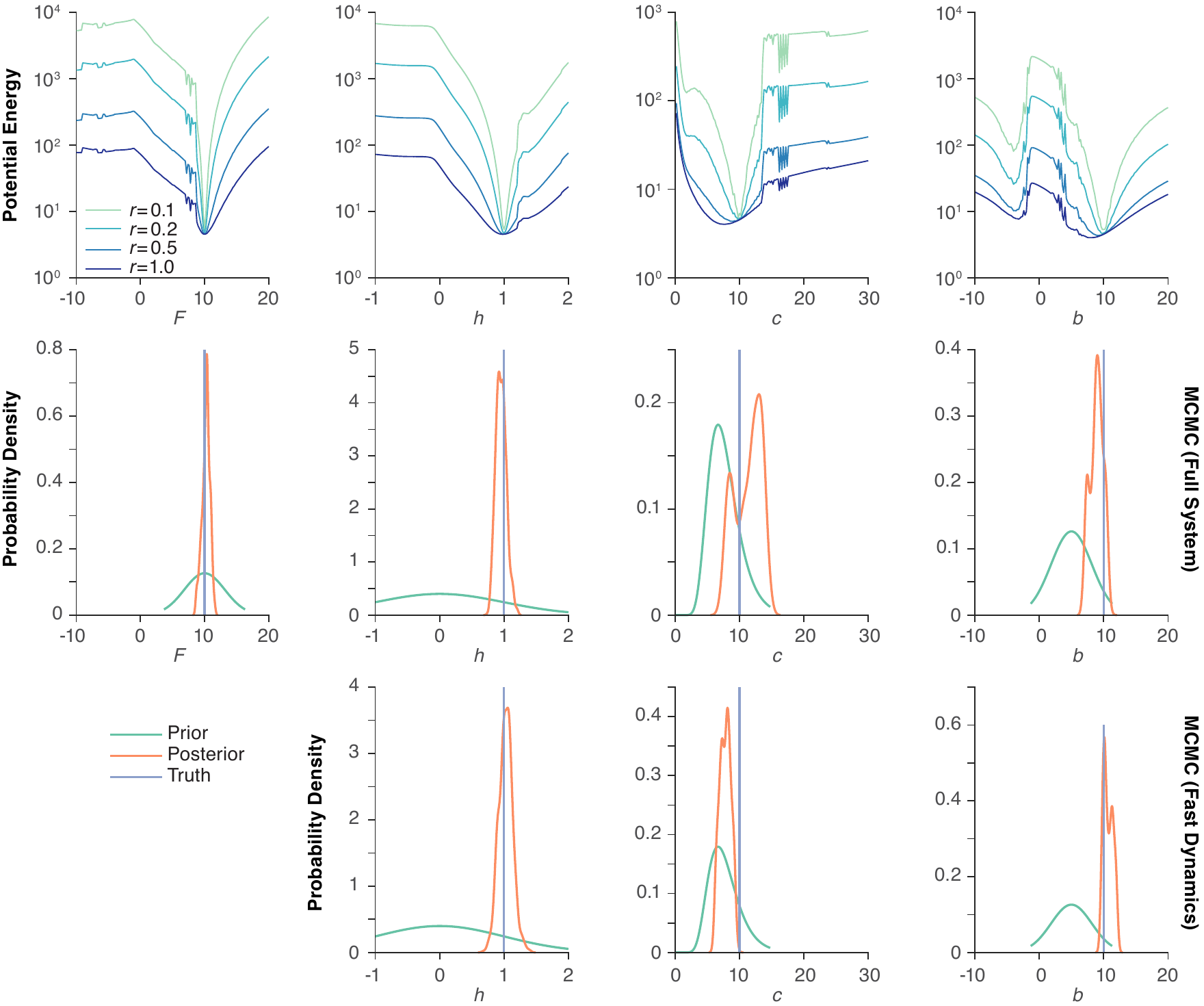}
   \caption{Probability density functions for parameter learning in Lorenz-96 model. First row: Marginal potential energies (negative logarithm of posterior PDF) at different noise levels $r$. Each marginal potential energy is obtained by varying the parameter on the abscissa, while holding other parameters fixed at their true values. Second row: Marginal prior and posterior PDFs, estimated by MCMC from the full dynamics \eqref{e:l96_slow}--\eqref{e:l96_fast},  with noise level $r=0.5$ and accumulating over $T=100~\mathrm{days}$ for each sample. Third row: Marginal prior and posterior PDFs, estimated by MCMC from the fast dynamics \eqref{e:l96_fast} only, likewise with noise level $r'=0.5$ but accumulating over only $T'=20~\mathrm{days}$ for each sample.}
   \label{fig:lorenz96_mcmc}
\end{figure*}
To illustrate the landscape learning algorithms have to navigate, Figure~\ref{fig:lorenz96_mcmc} (top row) shows sections through the potential energy, defined as the negative logarithm of the posterior PDF,
\begin{equation}
U(\vec{\theta}) = J_o(\vec{\theta}) - \sum_{i=1}^4 \log p_i(\theta_i),
\end{equation}
where $p_i(\theta_i)$ is the prior PDF of parameter $\theta_i$. The figure shows the marginal potential energies obtained as one parameter at a time is varied and the objective function $J_o(\vec{\theta})$ is accumulated by forward integration, while the other parameters are held fixed at their true values. As the noise level $r$ increases, the contribution of the log-likelihood of the data ($\propto J_o$) is down-weighted relative to the prior, the posterior modes shift toward the prior modes, and the posterior is smoothed. Here the objective function $J_o(\vec{\theta})$ for each parameter setting is accumulated over a long period ($10^4$~days) to minimize sampling variability. However, even with this wide accumulation window, sampling variability remains in some parameter regimes and there noticeably affects $J_o(\vec{\theta})$. An example is the roughness around $c=17$, which appears to be caused by metastability on timescales longer than the accumulation window.  The roughness could be smoothed by accumulating over periods that are yet longer, or by averaging over an ensemble of initial conditions, but analogous smoothing might be impractical for ESMs. Time-averaged ESM statistics may exhibit similarly rough dependencies on some parameters \citep[e.g.,][]{Suzuki13a,Zhao16a}, although the dependence on other parameters appears to be relatively smooth \citep[e.g.,][]{Neelin10a}, perhaps because ESM parameters targeted for tuning are chosen for the smooth dependence of the climate state on them. Roughness of the potential energy landscape can present challenges for learning algorithms, which may get stuck in local minima. Note also the bimodality in $b$, which arises because the one-point statistics we fit cannot easily distinguish prograde wave modes of the system (which propagate toward increasing $k$) from retrograde modes \citep[cf.][]{Lorenz98a}.

\subsubsection{Bayesian Inversion}

We use the random-walk Metropolis (RWM) MCMC algorithm \citep{Brooks11a} for a full Bayesian inversion of parameters in the dynamical system \eqref{e:l96_slow}, \eqref{e:l96_fast}, thereby sampling from the posterior PDF. To reduce burn-in (MCMC spin-up) time, we initialize the algorithm close to the true parameter values with the result of an ensemble Kalman inversion (see below). The RWM algorithm is then run over $2200$ iterations, the first $200$ iterations are discarded as burn-in, and the posterior PDF is estimated by binning every other of the remaining 2000 samples. The objective function for each sample is accumulated over $T=100~\mathrm{days}$, using the end state of the previous forward integration as initial condition for the next one, without discarding any spin-up after a parameter update. 

The resulting marginal posterior PDFs do not all peak exactly at the true parameter values, but the true parameter values lie in a region that contains most of the posterior probability mass (Figure~\ref{fig:lorenz96_mcmc}, second row). The posterior PDFs indicate the uncertainties inherent in estimating the parameters. The posterior PDF of $c$ has the largest spread, in terms of standard deviation normalized by mean, indicating relatively large uncertainty in this parameter. The uncertainty appears to arise from the roughness of the potential energy (Figure~\ref{fig:lorenz96_mcmc}, first row), which reflects inherent sensitivity of the system response to parameter variability; additional roughness of the posterior PDFs may be caused by sampling variability from finite-time  averages \citep{Wang14c}. For all four parameters, the posterior PDFs differ significantly from the priors, demonstrating the information content provided by the synthetic data. Finally, although these results have been obtained with $O(10^3)$ forward model integrations and objective function evaluations, more objective function evaluations may be required for more complex forward models, such as ESMs. 

\subsubsection{Ensemble Kalman Inversion}

Ensemble Kalman inversion may be an attractive learning algorithm for ESMs when Bayesian inversion with MCMC is computationally too demanding. To illustrate its performance, we use the algorithm of \citet{Iglesias13a}, initializing ensembles of size $M$ with parameters drawn from the prior PDFs. In the analysis step of the Kalman inversion, we perturb the target data by addition of noise with zero mean and variance given by \eqref{e:noise}, that is, replacing $\langle \vec{f}(\tilde X, \tilde Y) \rangle_\infty$ by $\langle \vec{f}(\tilde X, \tilde Y) \rangle_\infty + \vec{\eta}^{(j)}$ with $\vec{\eta}^{(j)}\sim \mathcal N(0,\Sigma)$ for each ensemble member $j$. As in the MCMC algorithm, the objective function for each parameter setting is accumulated over $T=100~\mathrm{days}$, without discarding any spin-up after each parameter update. As initial state for the integration of the ensemble, we use a state drawn from the statistically steady state of a simulation with the true parameters.

\begin{figure}[htb] %  figure placement: here, top, bottom, or page
   \centering
   \includegraphics{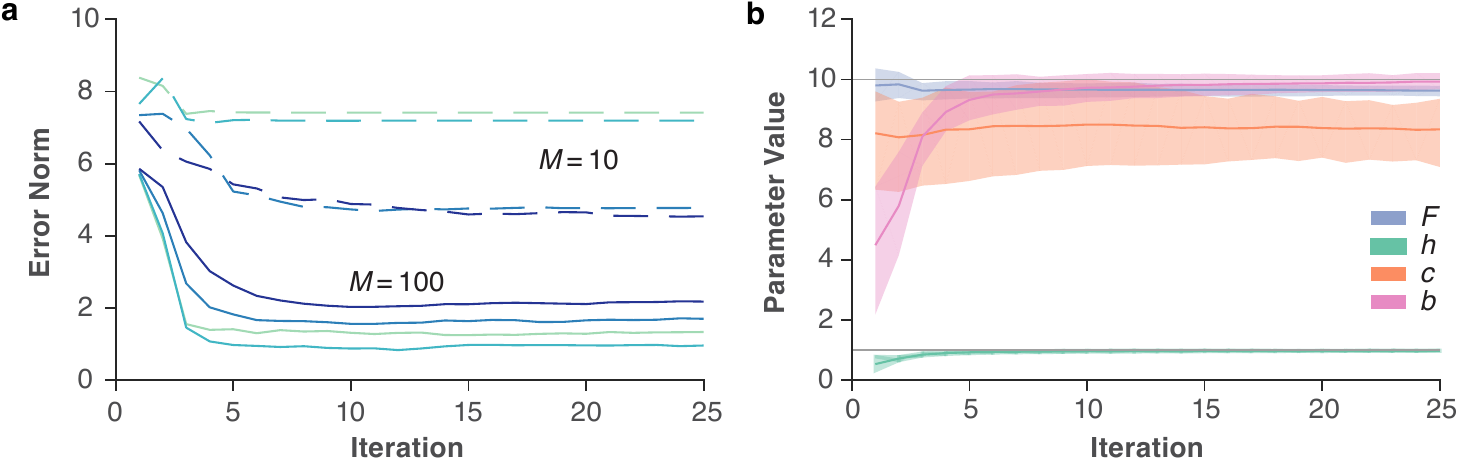}
   \caption{Convergence of ensemble Kalman inversion. (a) Error norm $\|\vec{\hat\theta} - \vec{\tilde\theta}\|$ of ensemble-mean parameter estimate $\vec{\hat\theta}$ as a function of iteration, for different ensemble sizes $M$ and noise levels $r$. Dashed lines for $M=10$, and solid lines for $M=100$. Color coding of noise levels $r=0.1$, $0.2$, $0.5$, and $1.0$ as in Figure~\ref{fig:lorenz96_mcmc} (top row). (b) Ensemble-mean value $\hat\theta_i$ of the four parameters as a function of iteration (solid lines), with interquartile range of ensemble (shading from 25th to 75th percentile), for $M=100$ and $r=0.5$. The objective function for each parameter setting is accumulated over $T=100~\mathrm{days}$.}
   \label{fig:EnKFerr_lorenz96ms}
\end{figure}

\begin{table}[ht]
\centering
\begin{tabular}{l|ccc}
  \hline
\multicolumn{1}{c|}{Noise} & Mean ($M=10$) & Mean ($M=100$) & Std ($M=100$) \\ 
  \hline
% $r=0.1$ & (10.11, 0.939,  11.02, 8.45) & (9.82, 0.996, 9.19, 9.95) & (0.08, 0.002,  0.38, 0.04) \\ 
% $r=0.2$ & (10.23, 0.941, 11.34, 8.38) & (9.88, 0.994,  9.40, 9.93) & (0.22, 0.006,  1.01, 0.13) \\ 
% $r=0.5$ & (10.29, 0.964,	 11.70, 8.79) & (9.90, 0.984,  9.48, 9.78) & (0.34, 0.018, 1.63, 0.34) \\
% $r=1.0$ & (10.06, 0.949, 10.93, 8.55) & (9.86, 0.954,  9.31, 9.29) & (0.42, 0.033, 1.98, 0.60) \\
% %  $r=1.0$ (no perturb) & (10.35, 0.908,  11.58, 8.32) & (9.82, 0.917,  8.78, 8.87) & (0.38, 0.026, 1.98, 0.47) \\
$r=0.1$ & (9.62, 0.579, 9.37, 2.63) & (9.71, 0.992, 8.70, 9.95) & (0.023, 0.001, 0.104, 0.022) \\ 
$r=0.2$ & (9.57, 0.516, 7.90, 3.15) & (9.77, 0.994, 9.07, 10.04) & (0.107, 0.005, 0.524, 0.103) \\ 
$r=0.5$ & (9.77, 0.522, 9.29, 5.31) & (9.63, 0.982, 8.34, 9.93) & (0.295, 0.017, 1.477, 0.350) \\
%$r=0.5$ (no perturb) & (9.49, 0.513, 8.05, 4.46) & (9.76, 0.952, 8.89, 9.40) & (0.230, 0.011, 1.158, 0.216) \\
$r=1.0$ & (9.70, 0.633, 7.68, 6.13) & (9.53, 0.952, 7.97, 9.37) & (0.385, 0.039, 1.964, 0.701) \\
   \hline
\end{tabular}
\caption{Ensemble means and standard deviations for the parameters $\theta=(F, h, c, b)$ obtained by ensemble Kalman inversions for different ensemble sizes $M$ and different noise levels $r$.} 
\label{tab:EnKF_opt}
\end{table}
Table~\ref{tab:EnKF_opt} summarizes the solutions obtained by this ensemble Kalman inversion after $N_{\max}=25$ iterations, for different ensemble sizes $M$ and noise levels $r$. The ensemble mean of the Kalman inversion provides reasonable parameter estimates. But the ensemble standard deviation does not always provide quantitatively accurate uncertainty information. For example, for low noise levels, the true parameter values often lie more than two standard deviations away from the ensemble mean. The ensemble spread also differs quantitatively from the posterior spread in the MCMC simulations. In experiments in which we did not perturb the target data, the smaller ensembles ($M=10$) occasionally collapsed, with each ensemble member giving the same point estimate of the parameters. In such cases, the ensemble contains no uncertainty information, illustrating potential pitfalls of using ensemble Kalman inversion for uncertainty quantification. However, with the perturbed data and for larger ensembles, the ensemble standard deviation is qualitatively consistent with the posterior PDF estimated by MCMC (Figure~\ref{fig:lorenz96_mcmc}, second row). It provides some uncertainty information, especially for higher noise levels, for example, in the sense that the parameter $c$ is demonstrably the most uncertain (Table~\ref{tab:EnKF_opt} and Figure~\ref{fig:EnKFerr_lorenz96ms}b). Methods such as localization and variance inflation can help with issues related to ensemble collapse and can also be used to improve ensemble statistics more generally (see \cite{Law15a} and the references therein). However, systematic principles for their application with the aim of correctly reproducing Bayesian posterior statistics have not been found, and so we have not adopted this approach.

The ensemble Kalman inversion typically converges within a few iterations (Figure~\ref{fig:EnKFerr_lorenz96ms} indicates ${\lesssim}\,5$ iterations when $M=100$). Larger ensembles lead to solutions closer to the truth (Figure~\ref{fig:EnKFerr_lorenz96ms}a). Convergence within 5 iterations for ensembles of size 10 or 100 implies 50 or 500 objective function evaluations, representing substantial computational savings over the MCMC algorithm with 2000 objective function evaluations. These computational savings come at the expense of detailed uncertainty information. Where the optimal trade-off lies between computational efficiency, on the one hand, and precision of parameter estimates and uncertainty quantification, on the other hand, remains to be investigated. 

\subsection{Parameter Learning From Fast Dynamics}

Finally, we investigate learning about parameters from the fast dynamics \eqref{e:l96_fast} alone. This is similar to learning about computable parameters from local high-resolution simulations, e.g., of clouds. That is, the fast dynamics \eqref{e:l96_fast} with the true parameters $\vec{\tilde\theta}$ stand for the high-resolution model $\cL$, which generates data $\vec{\tilde z} = \tilde Y$, and the fast dynamics with parameters $\vec{\theta}$ play the role of the single-column model $\cS$, which generates data $\vec{z} = Y$. We choose $k=1$ and fix $X_1=2.556$, a value taken from the statistically steady state of the full dynamics. There are three parameters to learn from the fast dynamics: $(\theta_2, \theta_3, \theta_4) = (h,c,b)$. The one-point statistics $(\bar Y_1,\overline{Y_1^2})$ of the fast variables are not enough to recover all three. Therefore, we consider the moment function 
\begin{equation}
\vec{g}(Y) = \left( 
\begin{array}{c} 
Y_{j,1}\\
Y_{j,1}Y_{j',1}
\end{array}
\right), \qquad j, j' \in \{1, \dots, J\},
\end{equation}
containing all first moments and second moments (including cross-moments), giving a vector of length  $J+J(J+1)/2=65$. We minimize the ``high-resolution'' objective function 
\begin{equation}\label{e:obj_o2}
J_s(\theta_2, \theta_3, \theta_4)=\frac12\| \langle \vec{g}(Y)  \rangle_{T'} - \langle \vec{g}(\tilde Y)
\rangle_{\infty} \|_{\Sigma'}^2,
\end{equation}
using a diagonal noise covariance matrix $\Sigma'$ with diagonal elements proportional to the variances of the statistics in $\vec{g}$, with a noise level $r'$  analogous to the noise covariance matrix \eqref{e:noise}. The variances of the statistics are estimated from a long control integration of the fast dynamics with fixed $X_1 = 2.556$. Because the fast variables $Y$ evolve more rapidly than the slow variables $X$, we accumulate statistics over only $T'=20~\mathrm{days}$. 

Bayesian inversion with RWM, with the same priors and algorithmic settings as before and with noise level $r'=0.5$, again gives marginal posterior PDFs with modes close to the truth (Figure~\ref{fig:lorenz96_mcmc}, third row). The posterior PDFs exhibit similar multi-modality and reflect similar uncertainties and biases of posterior modes as those obtained from the full dynamics, especially with respect to the relatively large uncertainties in $c$ (cf.\ Figure~\ref{fig:lorenz96_mcmc}, second row). 

These examples illustrate the potential of learning about parameters from observations and from local high-resolution simulations under selected conditions (here, for just one value of the slow variable $X_1$). An important question for future investigations is to what extent such results generalize to imperfect parameterization schemes, whose dynamics is usually not identical to the data-generating dynamics, so that structural in addition to parametric uncertainties arise. This issue can be studied for the Lorenz-96 system, for example, by using approximate models as parameterizations of the fast dynamics \citep[e.g.,][]{Fatkullin04a,Wilks05a,Crommelin08a}.

% In order to understand the role of imperfect parameterization schemes for fast variables, such as those used
% to represent clouds, we will also study the use of the model system 
% \begin{align}
% \frac{dX_k}{dt}     &= - X_{k-1}(X_{k-2} - X_{k+1}) - X_k + F - h \bar{Y}_k,\\
% \frac{d\bar{Y}_{k}}{dt} &= P(X_k,\bar{Y}_{k})+\sqrt{2\beta}\eta_k
% \end{align}
% where $\eta_k$ is a temporally white noise process. In ths context $P$ will be a polynomial
% function, and it is of interest to train parameters in $P$, and the parameter $\beta$, 
% to data generated using  fully resolved simulations of the original Lorenz model. 

% interaction of high-resolution simulations, learning parameters in parameterization, and slow simulations: uniformly sample $k$'s to high-res simulations?

\section{Outlook}\label{s:discussion}

Just as weather forecasts have made great strides over the past decades thanks to improvements in the assimilation of observations \citep{Bauer15a}, climate projections can advance similarly by harnessing observations and modern computational capabilities more systematically. New methods from data assimilation, inverse problems, and machine learning make it possible to integrate observations and targeted high-resolution simulations in an ESM that learns from both and uses both to quantify uncertainties. As an objective of such parameter learning we propose the reduction of biases and exploitation of emergent constraints through the matching of mean values and covariance components between ESMs, observations, and targeted high-resolution simulations. 

Coordinated space-based observations of crucial processes in the climate system are now available. For example, more than a decade's worth of coordinated observations of clouds, precipitation, temperature, and humidity with global coverage is available; parameterizations of clouds, convection, and turbulence can learn from them. Or, simultaneous measurements of \coo\ concentrations and photosynthesis are becoming available; parameterizations of terrestrial ecosystems can learn from them. So far, such observations have been primarily used to evaluate models and identify their deficiencies. Their potential to improve models has not yet been harnessed. Additionally, it is feasible to conduct faithful local high-resolution simulations of processes such as the dynamics of clouds or sea ice, which are in principle computable but are too costly to compute globally. Parameterizations can also learn from such high-resolution simulations, either online by nesting them in an ESM or offline by creating libraries of high-resolution simulations representing different regions and climates to learn from. Such a systematic approach to learning parameterizations from data allows the quantification of uncertainties in parameterizations, which in turn can be used to produce ensembles of climate simulations to quantify the uncertainty in predictions.

The machine learning of parameterizations in our view should be informed by the governing equations of subgrid-scale processes whenever they are known. The governing equations can be systematically coarse-grained, for example, by modeling the joint PDF of the relevant variables as a mixture of Gaussian kernels and generating moment equations for the modeled PDF from the governing equations \citep[cf.][]{Lappen01a,Golaz02a,Guo15a,Firl15a}. The closure parameters that necessarily arise in any such coarse-graining of nonlinear governing equations can then be learned from a broad range of observations and high-resolution simulations, as parametric or nonparametric functions of ESM state variables \citep[cf.][]{Parish16a}. The fineness of the coarse-graining (measured by the number of Gaussian kernels in the above example) can adapt to the information available to learn closure parameters. Such equation-informed machine learning will provide a more versatile means of modeling subgrid-scale processes than the traditional approach of fixing closure parameters ad hoc or on the basis of a small sample of observations or high-resolution simulations. Because parameterizations learned within the structure of the known governing equations respect the relevant symmetries and conservation laws to within the closure approximations, they likely have greater out-of-sample predictive power than unstructured parameterization schemes, such as neural networks that are fit to subgrid-scale processes without explicit regard for symmetries  and conservation laws \citep[e.g.,][]{Krasnopolsky13a}. Out-of-sample predictive power will be crucial if high-resolution simulations performed in selected locations and under selected conditions are to provide information globally and in changed climates. However, for non-computable processes whose governing equations are unknown, like many ecological or biogeochemical processes, more empirical, data-driven parameterization approaches may well be called for. 

An ESM that is designed from the outset to learn systematically from observations and high-resolution simulations represents an opportunity to achieve a leap in fidelity of parameterization schemes and thus of climate projections. Such an ESM can be expected to have attendant benefits for weather forecasting, because weather forecasting models and the atmospheric component of ESMs are essentially the same. However, challenges lie along the path toward realization of such an ESM:
\begin{itemize}
\item We need innovation in learning algorithms. Our relatively simple example showed that parameters in a perfect-model setting can be learned effectively and efficiently by ensemble Kalman inversion. It remains to investigate questions such as the optimal ensemble size in Kalman inversions, how to adapt inversion algorithms to imperfect models, and how to quantify uncertainties. To increase computational efficiency, online filtering algorithms need to be developed that update parameters on the fly as Earth system statistics are being accumulated.
\item We need investigations of the best metrics to use when  learning parameterization schemes from observations or high-resolution simulations. For example, are least-squares objective functions the best ones to use? Which covariance components or other statistics should be included in the objective functions? There are  trade-offs between the number of covariance components that can be estimated from data and the information they can provide about parameterization schemes. 
\item We need innovation in how learning from observations should interact with learning from targeted high-resolution simulations. How should high-resolution simulations be targeted? Where is the optimum trade-off between the added computational cost of conducting high-resolution simulations and the marginal information about parameterization schemes they provide?
\item We need innovation in parameterization schemes themselves, to design them such that they can learn effectively from diverse data sources and can be systematically refined when more information becomes available. It will be important to develop parameterizations that treat subgrid-scale motions (e.g., boundary layer turbulence, shallow convection, deep convection) in a unified manner, to eliminate artificial spectral gaps that do not exist in nature and to reduce the number of correlated parameters in the schemes \citep[e.g.,][]{Lappen01a, Lappen01b,Kohler11a,Suselj13a,Park14a,Park14b,Guo15a}. Novel approaches that exploit ideas ranging from stochastic parameterization to systematic coarse-graining likely have roles to play here \citep[e.g.,][]{Majda03a,Majda08a,Klein06a,Palmer05a,Palmer10a,Majda12a,Wouters13a,Lucarini14a,Wouters16b,Berner17a}. Furthermore, as the resolution of ESMs increases, it will also be necessary to revisit the common practice of modeling subgrid-scale dynamics in grid columns, because the lateral exchange of subgrid-scale information across grid columns will play increasingly important roles.
\end{itemize}

The time is right to seize the opportunities that the available global observations and our computational resources present. Fundamentally re-engineering atmospheric parameterization schemes, such as cloud and boundary layer parameterizations, will become a necessity as atmosphere models, within the next decade, reach horizontal grid spacings of 1--10~km and begin to resolve deep convection \citep{Schneider17a}. At such resolutions, common assumptions made in existing parameterization schemes, such as that clouds and the planetary boundary layer adjust instantaneously to changes in resolved-scale dynamics, will become untenable. Additionally, advances in high-performance computing (e.g., many-core computational architectures based on graphical processing units) will soon require a re-design of the software infrastructure of ESMs \citep{Bretherton12a,Schulthess15a,Schalkwijk15a}. So it is timely now to re-engineer ESMs and parameterization schemes, and design them from the outset so that they can learn systematically from observations and targeted high-resolution simulations. 

\begin{figure}[h]
\centering
\includegraphics[width=\textwidth]{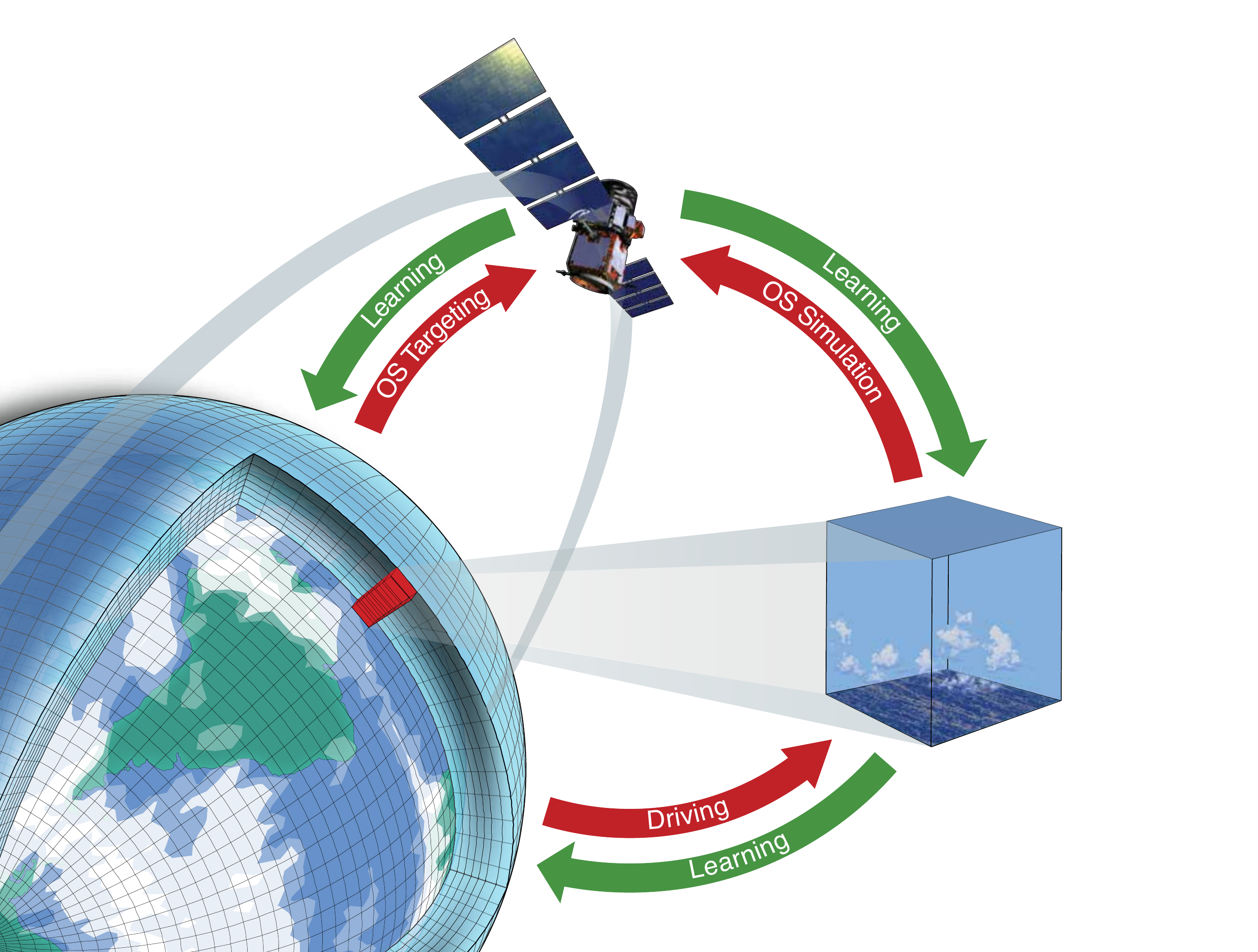}
\caption{Schematic of Earth system modeling framework that integrates global observing systems (OS) and targeted high-resolution simulations. Not only can parameterizations in ESMs and high-resolution simulations learn from observations. The same framework and the observing system simulators that it needs to encompass can also be used for observing system simulation experiments to assess the value and benefits of new observing systems. (Image credit for satellite: NASA.)}
\label{f:schematic}
\end{figure}
Integrating observations and targeted high-resolution simulations in an Earth system modeling framework would have multiple attendant benefits. Solving the inverse problems of learning about parameterizations from observations requires observing system simulators that map model state variables to observables (Figure~\ref{f:schematic}). The same observing system simulators, integrated in an Earth system modeling framework, can be used to answer questions about the value new observations would provide, for example, in terms of reduced uncertainties in ESMs. Addressing such questions in observing system simulation experiments (OSSEs) is increasingly required before the acquisition of new observing systems (e.g., as part of the U.S.\ Weather Research and Forecasting Innovation Act of 2017). They are naturally answered within the framework we propose.

\acknowledgments
We gratefully acknowledge financial support by Charles Trimble, by the Office Naval Research (grant N00014-17-1-2079), and by the President's and Director's Fund of Caltech and the Jet Propulsion Laboratory. We also thank V.\ Balaji, Michael Keller, Dan McCleese, and John Worden for helpful discussions and comments on drafts, and Momme Hell for preparing Figure~\ref{f:schematic}. The program code used in this paper is available at climate-dynamics.org/publications/. Part of this research was carried out at the Jet Propulsion Laboratory, California Institute of Technology, under a contract with the National Aeronautics and Space Administration.


\begin{thebibliography}{196}
\providecommand{\natexlab}[1]{#1}
\expandafter\ifx\csname urlstyle\endcsname\relax
  \providecommand{\doi}[1]{doi:\discretionary{}{}{}#1}\else
  \providecommand{\doi}{doi:\discretionary{}{}{}\begingroup
  \urlstyle{rm}\Url}\fi

\bibitem[{\textit{Adam et~al.}(2016)\textit{Adam, Schneider, Brient, and
  Bischoff}}]{Adam16b}
Adam, O., T.~Schneider, F.~Brient, and T.~Bischoff (2016), Relation of the
  double-{ITCZ} bias to the atmospheric energy budget in climate models,
  \textit{Geophys. Res. Lett.}, \textit{43}, 7670--7677,
  \doi{10.1002/2016GL069465}.

\bibitem[{\textit{Adam et~al.}(2017)\textit{Adam, Schneider, and
  Brient}}]{Adam17a}
Adam, O., T.~Schneider, and F.~Brient (2017), Regional and seasonal variations
  of the double-{ITCZ} bias in {CMIP5} models, \textit{Climate Dyn.},
  \doi{10.1007/s00382-017-3909-1}.

\bibitem[{\textit{Aksoy et~al.}(2006)\textit{Aksoy, Zhang, and
  Nielsen-Gammon}}]{Aksoy06a}
Aksoy, A., F.~Zhang, and J.~W. Nielsen-Gammon (2006), Ensemble-based
  simultaneous state and parameter estimation with {MM5}, \textit{Geophys. Res.
  Lett.}, \textit{33}, L12,801, \doi{10.1029/2006GL026186}.

\bibitem[{\textit{Alexanderian et~al.}(2016)\textit{Alexanderian, Petra,
  Stadler, and Ghattas}}]{Alexanderian16a}
Alexanderian, A., N.~Petra, G.~Stadler, and O.~Ghattas (2016), A fast and
  scalable method for {A}-optimal design of experiments for
  infinite-dimensional {B}ayesian nonlinear inverse problems, \textit{SIAM J.
  Sci. Comp.}, \textit{38}, A243--A272.

\bibitem[{\textit{Anderson et~al.}(2009)\textit{Anderson, Hoar, Raeder, Liu,
  Collins, Torn, and Avellano}}]{Anderson09a}
Anderson, J., T.~Hoar, K.~Raeder, H.~Liu, N.~Collins, R.~Torn, and A.~Avellano
  (2009), The data assimilation research testbed: {A} community facility,
  \textit{Bull. Amer. Meteor. Soc.}, \textit{90}, 1283--1296,
  \doi{10.1175/2009BAMS2618.1}.

\bibitem[{\textit{Anderson}(2001)}]{Anderson01a}
Anderson, J.~L. (2001), An ensemble adjustment {K}alman filter for data
  assimilation, \textit{Mon. Wea. Rev.}, \textit{129}, 2884--2903,
  \doi{10.1175/1520-0493(2001)129<2884:AEAKFF>2.0.CO;2}.

\bibitem[{\textit{Annan and Hargreaves}(2007)}]{Annan07a}
Annan, J.~D., and J.~C. Hargreaves (2007), Efficient estimation and ensemble
  generation in climate modelling, \textit{Phil. Trans. R. Soc. A},
  \textit{365}, 2077--2088, \doi{10.1098/rsta.2007.2067}.

\bibitem[{\textit{Ban et~al.}(2015)\textit{Ban, Schmidli, and
  Sch{\"a}r}}]{Ban15a}
Ban, N., J.~Schmidli, and C.~Sch{\"a}r (2015), Heavy precipitation in a
  changing climate: Does short-term summer precipitation increase faster?,
  \textit{Geophys. Res. Lett.}, \textit{42}, 1165--1172,
  \doi{10.1002/2014GL062588}.

\bibitem[{\textit{Bauer et~al.}(2015)\textit{Bauer, Thorpe, and
  Brunet}}]{Bauer15a}
Bauer, P., A.~Thorpe, and G.~Brunet (2015), The quiet revolution of numerical
  weather prediction, \textit{Nature}, \textit{525}, 47--55,
  \doi{10.1038/nature14956}.

\bibitem[{\textit{Benedict and Randall}(2009)}]{Benedict09a}
Benedict, J.~J., and D.~A. Randall (2009), Structure of the {M}adden-{J}ulian
  oscillation in the superparameterized {CAM}, \textit{J. Atmos. Sci.},
  \textit{66}, 3277--3296, \doi{10.1175/2009JAS3030.1}.

\bibitem[{\textit{Berner et~al.}(2017)\textit{Berner, Achatz, Batte, Bengtsson,
  De~La~Camara, Christensen, Colangeli, Coleman, Crommelin, Dolaptchiev
  et~al.}}]{Berner17a}
Berner, J., U.~Achatz, L.~Batte, L.~Bengtsson, A.~De~La~Camara, H.~M.
  Christensen, M.~Colangeli, D.~R. Coleman, D.~Crommelin, S.~I. Dolaptchiev,
  et~al. (2017), Stochastic parameterization: towards a new view of weather and
  climate models, \textit{Bull. Am. Meteor. Soc.}, \textit{98}, 565--587,
  \doi{10.1175/BAMS-D-15-00268.1}.

\bibitem[{\textit{Beskos et~al.}(2017)\textit{Beskos, Girolami, Lan, Farrell,
  and Stuart}}]{Beskos17a}
Beskos, A., M.~Girolami, S.~Lan, P.~E. Farrell, and A.~M. Stuart (2017),
  Geometric {MCMC} for infinite-dimensional inverse problems, \textit{J. Comp.
  Phys.}, \textit{335}, 327--351, \doi{10.1016/j.jcp.2016.12.041}.

\bibitem[{\textit{Bishop et~al.}(2001)\textit{Bishop, Etherton, and
  Majumdar}}]{Bishop01a}
Bishop, C.~H., B.~J. Etherton, and S.~J. Majumdar (2001), Adaptive sampling
  with the ensemble transform {K}alman filter. {Part~I}: Theoretical aspects,
  \textit{Mon. Wea. Rev.}, \textit{129}, 420--436,
  \doi{10.1175/1520-0493(2001)129<0420:ASWTET>2.0.CO;2}.

\bibitem[{\textit{Bloom et~al.}(2016)\textit{Bloom, Exbrayat, van~der Velde,
  Feng, and Williams}}]{Bloom16a}
Bloom, A.~A., J.-F. Exbrayat, I.~R. van~der Velde, L.~Feng, and M.~Williams
  (2016), The decadal state of the terrestrial carbon cycle: Global retrievals
  of terrestrial carbon allocation, pools, and residence times, \textit{Proc.
  Natl. Acad. Sci.}, \textit{113}, 1285--1290,
  \doi{doi:10.5194/essd-5-165-2013}.

\bibitem[{\textit{Bocquet and Sakov}(2013)}]{Bocquet13a}
Bocquet, M., and P.~Sakov (2013), Joint state and parameter estimation with an
  iterative ensemble {K}alman smoother, \textit{Nonlin. Processes Geophys.},
  \textit{20}, 803--818, \doi{10.5194/npg-20-803-2013}.

\bibitem[{\textit{Bocquet and Sakov}(2014)}]{Bocquet14a}
Bocquet, M., and P.~Sakov (2014), An iterative ensemble {K}alman smoother,
  \textit{Quart. J. Roy. Meteor. Soc.}, \textit{140}, 1521--1535,
  \doi{10.1002/qj.2236}.

\bibitem[{\textit{Bodas-Salcedo et~al.}(2014)\textit{Bodas-Salcedo, Williams,
  Ringer, Beau, Cole, Dufresne, Koshiro, Stevens, Wang, and
  Yokohata}}]{Bodas-Salcedo14a}
Bodas-Salcedo, A., K.~D. Williams, M.~A. Ringer, I.~Beau, J.~N.~S. Cole, J.-L.
  Dufresne, T.~Koshiro, B.~Stevens, Z.~Wang, and T.~Yokohata (2014), Origins of
  the solar radiation biases over the southern ocean in {CFMIP2} models,
  \textit{J. Climate}, \textit{27}, 41--56, \doi{10.1175/JCLI-D-13-00169.1}.

\bibitem[{\textit{Bony et~al.}(2006)\textit{Bony, Colman, Kattsov, Allan,
  Bretherton, Dufresne, Hall, Hallegatte, Holland, Ingram, Randall, Soden,
  Tselioudis, and Webb}}]{Bony06}
Bony, S., R.~Colman, V.~M. Kattsov, R.~P. Allan, C.~S. Bretherton, J.-L.
  Dufresne, A.~Hall, S.~Hallegatte, M.~M. Holland, W.~Ingram, D.~A. Randall,
  B.~J. Soden, G.~Tselioudis, and M.~J. Webb (2006), How well do we understand
  and evaluate climate change feedback processes?, \textit{J. Climate},
  \textit{19}, 3445--3482, \doi{10.1175/JCLI3819.1}.

\bibitem[{\textit{Bretherton et~al.}(2012)\textit{Bretherton, Balaji, Delworth,
  Dickinson, Edmonds, Famiglietti, Fung, Hack, Hurrell, Jacob, III, Leung,
  Marshall, Masloswski, Mearns, Rood, Smarr et~al.}}]{Bretherton12a}
Bretherton, C., V.~Balaji, T.~Delworth, R.~E. Dickinson, J.~A. Edmonds, J.~S.
  Famiglietti, I.~Fung, J.~S. Hack, J.~S. Hurrell, D.~J. Jacob, J.~L.~K. III,
  L.-Y.~R. Leung, S.~Marshall, W.~Masloswski, L.~O. Mearns, R.~B. Rood, L.~L.
  Smarr, et~al. (2012), \textit{A National Strategy for Advancing Climate
  Modeling}, The National Academies Press, Washington, D.C.

\bibitem[{\textit{Brient and Schneider}(2016)}]{Brient16b}
Brient, F., and T.~Schneider (2016), Constraints on climate sensitivity from
  space-based measurements of low-cloud reflection, \textit{J. Climate},
  \textit{29}, 5821--5835, \doi{10.1175/JCLI-D-15-0897.1}.

\bibitem[{\textit{Brooks et~al.}(2011)\textit{Brooks, Gelman, Jones, and
  Meng}}]{Brooks11a}
Brooks, S., A.~Gelman, G.~L. Jones, and X.-L. Meng (2011), \textit{Handbook of
  {M}arkov {C}hain {M}onte {C}arlo}, 619 pp., Chapman and Hall/CRC.

\bibitem[{\textit{Canadell et~al.}(2007)\textit{Canadell, Qu{\'e}r{\'e},
  Raupach, Field, Buitenhuis, Ciais, Conway, Gillett, Houghton, and
  Marland}}]{Canadell07a}
Canadell, J.~G., C.~L. Qu{\'e}r{\'e}, M.~R. Raupach, C.~B. Field, E.~T.
  Buitenhuis, P.~Ciais, T.~J. Conway, N.~P. Gillett, R.~A. Houghton, and
  G.~Marland (2007), Contributions to accelerating atmospheric
  {CO\textsubscript{2}} growth from economic activity, carbon intensity, and
  efficiency of natural sinks, \textit{Proc. Natl. Acad. Sci.}, \textit{104},
  18,866--18,870, \doi{10.1073 pnas.0702737104}.

\bibitem[{\textit{Carrassi et~al.}(2017)\textit{Carrassi, Bocquet, Hannart, and
  Ghil}}]{Carrassi17a}
Carrassi, A., M.~Bocquet, A.~Hannart, and M.~Ghil (2017), Estimating model
  evidence using data assimilation, \textit{Quart. J. Roy. Meteor. Soc.},
  \textit{143}, 866--880, \doi{10.1002/qj.2972}.

\bibitem[{\textit{Cess et~al.}(1989)\textit{Cess, Potter, Blanchet, Boer, Ghan,
  Kiehl, Le~Treut, Li, Liang, Mitchell et~al.}}]{Cess89a}
Cess, R.~D., G.~Potter, J.~Blanchet, G.~Boer, S.~Ghan, J.~Kiehl, H.~Le~Treut,
  Z.-X. Li, X.-Z. Liang, J.~Mitchell, et~al. (1989), Interpretation of
  cloud-climate feedback as produced by 14 atmospheric general circulation
  models, \textit{Science}, \textit{245}, 513--516.

\bibitem[{\textit{Cess et~al.}(1990)\textit{Cess, Potter, Blanchet, Boer,
  Del~Genio, D{\'e}qu{\'e}, Dymnikov, Galin, Gates, Ghan, Kiehl, Lacis,
  Le~Treut, Li, Liang, McAvaney, Meleshko, Mitchell, Morcrette, Randall, Rikus,
  Roeckner, Royer, Schlese, Sheinin, Slingo, Sokolov, Taylor, Washington,
  Wetherald, Yagai, and Zhang}}]{Cess90a}
Cess, R.~D., G.~L. Potter, J.~P. Blanchet, G.~J. Boer, A.~D. Del~Genio,
  M.~D{\'e}qu{\'e}, V.~Dymnikov, V.~Galin, W.~L. Gates, S.~J. Ghan, J.~T.
  Kiehl, A.~A. Lacis, H.~Le~Treut, Z.-X. Li, X.-Z. Liang, B.~J. McAvaney, V.~P.
  Meleshko, J.~F.~B. Mitchell, J.-J. Morcrette, D.~A. Randall, L.~Rikus,
  E.~Roeckner, J.~F. Royer, U.~Schlese, D.~A. Sheinin, A.~Slingo, A.~P.
  Sokolov, K.~E. Taylor, W.~M. Washington, R.~T. Wetherald, I.~Yagai, and M.-H.
  Zhang (1990), Intercomparison and interpretation of climate feedback
  processes in 19 atmospheric general circulation models, \textit{J. Geophys.
  Res.}, \textit{95}, 16,601--16,615, \doi{10.1029/JD095iD10p16601}.

\bibitem[{\textit{Collins et~al.}(2012)\textit{Collins, Chandler, Cox,
  Huthnance, Rougier, , and Stephenson}}]{Collins12a}
Collins, M., R.~E. Chandler, P.~M. Cox, J.~M. Huthnance, J.~Rougier, , and
  D.~B. Stephenson (2012), Quantifying future climate change, \textit{Nature
  Climate Change}, \textit{2}, 403--409, \doi{10.1038/NCLIMATE1414}.

\bibitem[{\textit{Cotter et~al.}(2013)\textit{Cotter, Roberts, Stuart, and
  White}}]{Cotter13a}
Cotter, S.~L., G.~O. Roberts, A.~M. Stuart, and D.~White (2013), {MCMC} methods
  for functions: {M}odifying old algorithms to make them faster,
  \textit{Statist. Science}, \textit{28}(3), 424--446.

\bibitem[{\textit{Cox et~al.}(2013)\textit{Cox, Pearson, Booth, Friedlingstein,
  Huntingford, Jones, and Luke}}]{Cox13a}
Cox, P.~M., D.~Pearson, B.~B. Booth, P.~Friedlingstein, C.~Huntingford, C.~D.
  Jones, and C.~M. Luke (2013), Sensitivity of tropical carbon to climate
  change constrained by carbon dioxide variability, \textit{Nature},
  \textit{494}, 341--344, \doi{10.1038/nature11882}.

\bibitem[{\textit{Crisp et~al.}(2004)\textit{Crisp, Atlas, Breon, Brown,
  Burrows, Ciais, Connor, Doney, Fung, Jacob, Miller et~al.}}]{Crisp04a}
Crisp, D., R.~M. Atlas, F.-M. Breon, L.~R. Brown, J.~P. Burrows, P.~Ciais,
  B.~J. Connor, S.~C. Doney, I.~Y. Fung, D.~J. Jacob, C.~E. Miller, et~al.
  (2004), The {O}rbiting {C}arbon {O}bservatory ({OCO}) mission, \textit{Adv.
  Space Res.}, \textit{34}, 700--709, \doi{10.1016/j.asr.2003.08.062}.

\bibitem[{\textit{Crommelin and Vanden-Eijnden}(2008)}]{Crommelin08a}
Crommelin, D., and E.~Vanden-Eijnden (2008), Subgrid-scale parameterization
  with conditional {M}arkov chains, \textit{J. Atmos. Sci.}, \textit{65},
  2661--2675, \doi{10.1175/2008JAS2566.1}.

\bibitem[{\textit{{de Rooy} et~al.}(2013)\textit{{de Rooy}, Bechtold,
  Fr{\"o}hlich, Hohenegger, Jonker, Mironov, Siebesma, Teixeira, and
  Yano}}]{de-Rooy13a}
{de Rooy}, W.~C., P.~Bechtold, K.~Fr{\"o}hlich, C.~Hohenegger, H.~Jonker,
  D.~Mironov, A.~P. Siebesma, J.~Teixeira, and J.-I. Yano (2013), Entrainment
  and detrainment in cumulus convection: an overview, \textit{Quart. J. Roy.
  Meteor. Soc.}, \textit{139}, 1--19.

\bibitem[{\textit{Dee}(1995)}]{Dee95a}
Dee, D.~P. (1995), On-line estimation of error covariance parameters for
  atmospheric data assimilation, \textit{Mon. Wea. Rev.}, \textit{123},
  1128--1145, \doi{10.1175/1520-0493(1995)123<1128:OLEOEC>2.0.CO;2}.

\bibitem[{\textit{Dee}(2005)}]{Dee05a}
Dee, D.~P. (2005), Bias and data assimilation, \textit{Quart. J. Roy. Meteor.
  Soc.}, \textit{131}, 3323--3343, \doi{10.1256/qj.05.137}.

\bibitem[{\textit{Del~Moral et~al.}(2006)\textit{Del~Moral, Doucet, and
  Jasra}}]{Del-Moral06a}
Del~Moral, P., A.~Doucet, and A.~Jasra (2006), Sequential {M}onte {C}arlo
  samplers, \textit{J. Roy. Statist. Soc. B}, \textit{68}, 411--436,
  \doi{10.1111/j.1467-9868.2006.00553.x}.

\bibitem[{\textit{DeMott et~al.}(2013)\textit{DeMott, Stan, and
  Randall}}]{DeMott13a}
DeMott, C.~A., C.~Stan, and D.~A. Randall (2013), Northward propagation
  mechanisms of the boreal summer intraseasonal oscillation in the
  {ERA}-{I}nterim and {SP-CCSM}, \textit{J. Climate}, \textit{26}, 1973--1992,
  \doi{10.1175/JCLI-D-12-00191.1}.

\bibitem[{\textit{Devenish et~al.}(2012)\textit{Devenish, Bartello, Brenguier,
  Collins, Grabowski, IJzermans, Malinowski, Reeks, Vassilicos, L.-P.Wang, and
  Z.Warhaft}}]{Devenish12a}
Devenish, B.~J., P.~Bartello, J.-L. Brenguier, L.~R. Collins, W.~W. Grabowski,
  R.~H.~A. IJzermans, S.~P. Malinowski, M.~W. Reeks, J.~C. Vassilicos,
  L.-P.Wang, and Z.Warhaft (2012), Droplet growth in warm turbulent clouds,
  \textit{Quart. J. Roy. Meteor. Soc.}, \textit{138}, 1401--1429,
  \doi{10.1002/qj.1897}.

\bibitem[{\textit{Draper}(1995)}]{Draper95}
Draper, D. (1995), Assessment and propagation of model uncertainty, \textit{J.
  Roy. Statist. Soc. B}, \textit{57}, 45--97.

\bibitem[{\textit{E et~al.}(2007)\textit{E, Engquist, Li, Ren, and
  Vanden-Eijnden}}]{E07a}
E, W., B.~Engquist, X.~Li, W.~Ren, and E.~Vanden-Eijnden (2007), Heterogeneous
  multiscale methods: {A} review, \textit{Commun. Comput. Phys.}, \textit{3},
  367--450.

\bibitem[{\textit{Eldering et~al.}(2017)\textit{Eldering, Wennberg, Crisp,
  Schimel, Gunson, Chatterjee, Liu, Schwandner, Sun, O'Dell, Frankenberg,
  Taylor, Fisher, Osterman, Wunch, Hakkarainen, Tamminen, and
  Weir}}]{Eldering17a}
Eldering, A., P.~O. Wennberg, D.~Crisp, D.~S. Schimel, M.~R. Gunson,
  A.~Chatterjee, J.~Liu, F.~M. Schwandner, Y.~Sun, C.~W. O'Dell,
  C.~Frankenberg, T.~Taylor, B.~Fisher, G.~B. Osterman, D.~Wunch,
  J.~Hakkarainen, J.~Tamminen, and B.~Weir (2017), The {O}rbiting {C}arbon
  {O}bservatory-2 early science investigations of regional carbon dioxide
  fluxes, \textit{Science}, \textit{358}, eaam5745,
  \doi{10.1126/science.aam5745}.

\bibitem[{\textit{Emanuel and \v{Z}ivkovi\'{c} Rothman}(1999)}]{Emanuel99a}
Emanuel, K.~A., and M.~\v{Z}ivkovi\'{c} Rothman (1999), Development and
  evaluation of a convection scheme for use in climate models, \textit{J.
  Atmos. Sci.}, \textit{56}, 1766--1782,
  \doi{10.1175/1520-0469(1999)056<1766:DAEOAC>2.0.CO;2}.

\bibitem[{\textit{Engl et~al.}(1996)\textit{Engl, Hanke, and
  Neubauer}}]{Engl96}
Engl, H.~W., M.~Hanke, and A.~Neubauer (1996), \textit{Regularization of
  Inverse Problems}, 321 pp., Kluwer Academic Publishers, Dordrecht.

\bibitem[{\textit{Fatkullin and Vanden-Eijnden}(2004)}]{Fatkullin04a}
Fatkullin, I., and E.~Vanden-Eijnden (2004), A computational strategy for
  multiscale systems with applications to lorenz 96 model, \textit{J. Comp.
  Phys.}, \textit{200}, 605--638, \doi{10.1016/j.jcp.2004.04.013}.

\bibitem[{\textit{Firl and Randall}(2015)}]{Firl15a}
Firl, G.~J., and D.~A. Randall (2015), Fitting and analyzing {LES} using
  multiple trivariate {G}aussians, \textit{J. Atmos. Sci.}, \textit{72},
  1094--1116, \doi{10.1175/JAS-D-14-0192.1}.

\bibitem[{\textit{Flato et~al.}(2013)\textit{Flato, Marotzke, Abiodun,
  Braconnot, Chou, Collins, Cox, Driouech, Emori, Eyring, Forest, Gleckler,
  Guilyardi, Jakob, Kattsov, Reason, and Rummukainen}}]{Flato13a}
Flato, G., J.~Marotzke, B.~Abiodun, P.~Braconnot, S.~C. Chou, W.~Collins,
  P.~Cox, F.~Driouech, S.~Emori, V.~Eyring, C.~Forest, P.~Gleckler,
  E.~Guilyardi, C.~Jakob, V.~Kattsov, C.~Reason, and M.~Rummukainen (2013),
  Evaluation of climate models, in \textit{Climate Change 2013: The Physical
  Science Basis. Contribution of Working Group I to the Fifth Assessment Report
  of the Intergovernmental Panel on Climate Change}, edited by T.~F. Stocker,
  D.~Qin, G.-K. Plattner, M.~Tignor, S.~K. Allen, J.~Boschung, A.~Nauels,
  Y.~Xia, V.~Bex, and P.~M. Midgley, chap.~9, pp. 741--853, Cambridge
  University Press, Cambridge, UK, and New York, NY, USA.

\bibitem[{\textit{Fox-Kemper et~al.}(2014)\textit{Fox-Kemper, Bachman, Pearson,
  and Reckinger}}]{Fox-Kemper14a}
Fox-Kemper, B., S.~Bachman, B.~Pearson, and S.~Reckinger (2014), Principles and
  advances in subgrid modelling for eddy-rich simulations, \textit{CLIVAR
  Exchanges No .65}, \textit{19}(2).

\bibitem[{\textit{Frankenberg et~al.}(2011)\textit{Frankenberg, Fisher, Worden,
  Badgley, Saatchi, Lee, Toon, Butz, Jung, Kuze, and Yokota}}]{Frankenberg11a}
Frankenberg, C., J.~B. Fisher, J.~Worden, G.~Badgley, S.~S. Saatchi, J.-E. Lee,
  G.~C. Toon, A.~Butz, M.~Jung, A.~Kuze, and T.~Yokota (2011), New global
  observations of the terrestrial carbon cycle from {GOSAT}: Patterns of plant
  fluorescence with gross primary productivity, \textit{Geophys. Res. Lett.},
  \textit{38}, L17,706, \doi{10.1029/2011GL048738}.

\bibitem[{\textit{Frankenberg et~al.}(2014)\textit{Frankenberg, O'Dell, Berry,
  Guanter, Joiner, K{\"o}hler, Pollock, and Taylor}}]{Frankenberg14a}
Frankenberg, C., C.~O'Dell, J.~Berry, L.~Guanter, J.~Joiner, P.~K{\"o}hler,
  R.~Pollock, and T.~E. Taylor (2014), Prospects for chlorophyll fluorescence
  remote sensing from the {O}rbiting {C}arbon {O}bservatory-2, \textit{Remote
  Sens. Env.}, \textit{147}, 1--12, \doi{10.1016/j.rse.2014.02.007}.

\bibitem[{\textit{Friedlingstein}(2015)}]{Friedlingstein15a}
Friedlingstein, P. (2015), Carbon cycle feedbacks and future climate change,
  \textit{Phil. Trans. R. Soc. A}, \textit{373}, 20140,421,
  \doi{10.1098/rsta.2014.0421}.

\bibitem[{\textit{Friedlingstein et~al.}(2006)\textit{Friedlingstein, Cox,
  Betts, Bopp, von Bloh, Brovkin, Cadule, Doney, Eby, Fung, Bala, John, Jones,
  Joos, Kato, Kawamiya, Knorr, Lindsay, Matthews, Raddatz, Rayner, Reick,
  Roeckner, Schnitzler, Schnur, Strassmann, , Weaver, Yoshikawa, and
  Zeng}}]{Friedlingstein06a}
Friedlingstein, P., P.~Cox, R.~Betts, L.~Bopp, W.~von Bloh, V.~Brovkin,
  P.~Cadule, S.~Doney, M.~Eby, I.~Fung, G.~Bala, J.~John, C.~Jones, F.~Joos,
  T.~Kato, M.~Kawamiya, W.~Knorr, K.~Lindsay, H.~D. Matthews, T.~Raddatz,
  P.~Rayner, C.~Reick, E.~Roeckner, K.-G. Schnitzler, R.~Schnur, K.~Strassmann,
  , A.~J. Weaver, C.~Yoshikawa, and N.~Zeng (2006), Climate--carbon cycle
  feedback analysis: Results from the {C\textsuperscript{4}MIP} model
  intercomparison, \textit{J. Climate}, \textit{19}, 3337--3353,
  \doi{10.1175/JCLI3800.1}.

\bibitem[{\textit{Friedlingstein et~al.}(2014)\textit{Friedlingstein,
  Meinshausen, Arora, Jones, Anav, Liddicoat, and Knutti}}]{Friedlingstein14a}
Friedlingstein, P., M.~Meinshausen, V.~K. Arora, C.~D. Jones, A.~Anav, S.~K.
  Liddicoat, and R.~Knutti (2014), Uncertainties in {CMIP5} climate projections
  due to carbon cycle feedbacks, \textit{J. Climate}, \textit{27}, 511--526,
  \doi{10.1175/JCLI-D-12-00579.1}.

\bibitem[{\textit{Friend et~al.}(2014)\textit{Friend, Lucht, Rademacher,
  Keribin, Betts, Cadule, Ciais, Clark, Dankers, Falloon et~al.}}]{Friend14a}
Friend, A.~D., W.~Lucht, T.~T. Rademacher, R.~Keribin, R.~Betts, P.~Cadule,
  P.~Ciais, D.~B. Clark, R.~Dankers, P.~D. Falloon, et~al. (2014), Carbon
  residence time dominates uncertainty in terrestrial vegetation responses to
  future climate and atmospheric {CO\textsubscript{2}}, \textit{Proc. Natl.
  Acad. Sci.}, \textit{111}, 3280--3285, \doi{10.1073/pnas.1222477110}.

\bibitem[{\textit{Gardner et~al.}(2013)\textit{Gardner, Moholdt, Cogley,
  Wouters, Arendt, Wahr, Berthier, Hock, Pfeffer, Kaser, Ligtenberg, Bolch,
  Sharp, Hagen, {van den Broeke}, and Paul}}]{Gardner13a}
Gardner, A.~S., G.~Moholdt, J.~G. Cogley, B.~Wouters, A.~A. Arendt, J.~Wahr,
  E.~Berthier, R.~Hock, W.~T. Pfeffer, G.~Kaser, S.~R.~M. Ligtenberg, T.~Bolch,
  M.~J. Sharp, J.~O. Hagen, M.~R. {van den Broeke}, and F.~Paul (2013), A
  reconciled estimate of glacier contributions to sea level rise: 2003 to 2009,
  \textit{Science}, \textit{340}, 852--857, \doi{10.1126/science.1234532}.

\bibitem[{\textit{Golaz et~al.}(2002)\textit{Golaz, Larson, and
  Cotton}}]{Golaz02a}
Golaz, J.-C., V.~E. Larson, and W.~R. Cotton (2002), A {PDF}-based model for
  boundary layer clouds. {Part I}: Method and model description, \textit{J.
  Atmos. Sci.}, \textit{59}, 3540--3551.

\bibitem[{\textit{Golaz et~al.}(2013)\textit{Golaz, Horowitz, and
  {II}}}]{Golaz13a}
Golaz, J.-C., L.~W. Horowitz, and H.~L. {II} (2013), Cloud tuning in a coupled
  climate model: Impact on 20th century warming, \textit{Geophys. Res. Lett.},
  \textit{40}, 2246--2251, \doi{10.1002/grl.50232}.

\bibitem[{\textit{Grabowski}(2001)}]{Grabowski01a}
Grabowski, W.~W. (2001), Coupling cloud processes with the large-scale dynamics
  using the cloud-resolving convection parameterization ({CRCP}), \textit{J.
  Atmos. Sci.}, \textit{58}, 978--997,
  \doi{10.1175/1520-0469(2001)058<0978:CCPWTL>2.0.CO;2}.

\bibitem[{\textit{Grabowski}(2016)}]{Grabowski16a}
Grabowski, W.~W. (2016), Towards global large eddy simulation:
  Super-parameterization revisited, \textit{J. Meteor. Soc. Japan},
  \textit{94}, 327--344, \doi{10.2151/jmsj.2016-017}.

\bibitem[{\textit{Grabowski and Smolarkiewicz}(1999)}]{Grabowski99a}
Grabowski, W.~W., and P.~K. Smolarkiewicz (1999), {CRCP}: A cloud resolving
  convection parameterization for modeling the tropical convecting atmosphere,
  \textit{Physica D}, \textit{133}, 171--178,
  \doi{10.1016/S0167-2789(99)00104-9}.

\bibitem[{\textit{Grabowski and Wang}(2013)}]{Grabowski13a}
Grabowski, W.~W., and L.-P. Wang (2013), Growth of cloud droplets in a
  turbulent environment, \textit{Ann. Rev. Fluid Mech.}, \textit{45}, 293--324,
  \doi{10.1146/annurev-fluid-011212-140750}.

\bibitem[{\textit{Grell and D{\'e}v{\'e}nyi}(2002)}]{Grell02a}
Grell, G.~A., and D.~D{\'e}v{\'e}nyi (2002), A generalized approach to
  parameterizing convection combining ensemble and data assimilation
  techniques, \textit{Geophys. Res. Lett..}, \textit{29}, 1693,
  \doi{10.1029/2002GL015311}.

\bibitem[{\textit{Guilyardi et~al.}(2009)\textit{Guilyardi, Wittenberg,
  Fedorov, Collins, Wang, Capotondi, van Oldenborgh~Oldenborgh, and
  Stockdale}}]{Guilyardi09a}
Guilyardi, E., A.~Wittenberg, A.~Fedorov, M.~Collins, C.~Wang, A.~Capotondi,
  G.~J. van Oldenborgh~Oldenborgh, and T.~Stockdale (2009), Understanding
  {El~Ni{\~n}o} in ocean-atmosphere general circulation models, \textit{Bull.
  Amer. Meteor. Soc.}, \textit{90}, 325--340, \doi{10.1175/2008BAMS2387.1}.

\bibitem[{\textit{Guo et~al.}(2015)\textit{Guo, Golaz, Donner, Wyman, Zhao, and
  Ginoux}}]{Guo15a}
Guo, H., J.-C. Golaz, L.~J. Donner, B.~Wyman, M.~Zhao, and P.~Ginoux (2015),
  {CLUBB} as a unified cloud parameterization: Opportunities and challenges,
  \textit{Geophys. Res. Lett.}, \textit{42}, 4540--4547,
  \doi{10.1002/2015GL063672}.

\bibitem[{\textit{Hall and Qu}(2006)}]{Hall06a}
Hall, A., and X.~Qu (2006), Using the current seasonal cycle to constrain snow
  albedo feedback in future climate change, \textit{Geophys. Res. Lett.},
  \textit{33}, L03,502, \doi{10.1029/2005GL025127}.

\bibitem[{\textit{Hohenegger and Bretherton}(2011)}]{Hohenegger11a}
Hohenegger, C., and C.~S. Bretherton (2011), Simulating deep convection with a
  shallow convection scheme, \textit{Atmos. Chem. Phys.}, \textit{11},
  10,389--10,406, \doi{10.5194/acp-11-10389-2011}.

\bibitem[{\textit{Holloway and Neelin}(2009)}]{Holloway09a}
Holloway, C.~E., and J.~D. Neelin (2009), Moisture vertical structure, column
  water vapor, and tropical deep convection, \textit{J. Atmos. Sci.},
  \textit{66}, 1665--1683.

\bibitem[{\textit{Hope}(2015)}]{Hope15a}
Hope, C. (2015), The \$10 trillion value of better information about the
  transient climate response, \textit{Phil. Trans. R. Soc. A}, \textit{373},
  20140,429, \doi{10.1098/rsta.2014.0429}.

\bibitem[{\textit{Hourdin et~al.}(2013)\textit{Hourdin, Grandpeix, Rio, Bony,
  Jam, Cheruy, Rochetin, Fairhead, Idelkadi, Musat, Dufresne, Lahellec,
  Lefebvre, and Roehrig}}]{Hourdin13a}
Hourdin, F., J.-Y. Grandpeix, C.~Rio, S.~Bony, A.~Jam, F.~Cheruy, N.~Rochetin,
  L.~Fairhead, A.~Idelkadi, I.~Musat, J.-L. Dufresne, A.~Lahellec, M.-P.
  Lefebvre, and R.~Roehrig (2013), {LMDZ5B}: the atmospheric component of the
  {IPSL} climate model with revisited parameterizations for clouds and
  convection, \textit{Clim. Dyn.}, \textit{40}, 2193--2222,
  \doi{10.1007/s00382-012-1343-y}.

\bibitem[{\textit{Hourdin et~al.}(2017)\textit{Hourdin, Mauritsen, Gettelman,
  Golaz, Balaji, Duan, Folini, Ji, Klocke, Qian et~al.}}]{Hourdin17a}
Hourdin, F., T.~Mauritsen, A.~Gettelman, J.-C. Golaz, V.~Balaji, Q.~Duan,
  D.~Folini, D.~Ji, D.~Klocke, Y.~Qian, et~al. (2017), The art and science of
  climate model tuning, \textit{Bull. Amer. Meteor. Soc.}, \textit{98},
  589--602, \doi{10.1175/BAMS-D-15-00135.1}.

\bibitem[{\textit{Houtekamer and Zhang}(2016)}]{Houtekamer16a}
Houtekamer, P.~L., and F.~Zhang (2016), Review of the ensemble {K}alman filter
  for atmospheric data assimilation, \textit{Mon. Wea. Rev.}, \textit{144},
  4489--4532, \doi{10.1175/MWR-D-15-0440.1}.

\bibitem[{\textit{Hung et~al.}(2013)\textit{Hung, Lin, Wang, Kim, Shinoda, and
  Weaver}}]{Hung13a}
Hung, M.-P., J.-L. Lin, W.~Wang, D.~Kim, T.~Shinoda, and S.~J. Weaver (2013),
  {MJO} and convectively coupled equatorial waves simulated by {CMIP5} climate
  models, \textit{J. Climate}, \textit{26}, 6185--6214,
  \doi{10.1175/JCLI-D-12-00541.1}.

\bibitem[{\textit{Iglesias}(2016)}]{Iglesias16a}
Iglesias, M.~A. (2016), A regularizing iterative ensemble {K}alman method for
  {PDE}-constrained inverse problems, \textit{Inverse Problems}, \textit{32},
  025,002, \doi{10.1088/0266-5611/32/2/025002}.

\bibitem[{\textit{Iglesias et~al.}(2013)\textit{Iglesias, Law, and
  Stuart}}]{Iglesias13a}
Iglesias, M.~A., K.~J.~H. Law, and A.~M. Stuart (2013), Ensemble {Kalman}
  methods for inverse problems, \textit{Inverse Problems}, \textit{29}, 045,001
  (20pp), \doi{10.1088/0266-5611/29/4/045001}.

\bibitem[{\textit{{Intergovernmental Panel on Climate Change}}(2013)}]{ipcc13}
{Intergovernmental Panel on Climate Change} (2013), \textit{Climate Change
  2013: {T}he Physical Science Basis}, Cambridge University Press, New York.

\bibitem[{\textit{Jackson et~al.}(2008)\textit{Jackson, Sen, Huerta, Deng, and
  Bowman}}]{Jackson08a}
Jackson, C.~S., M.~K. Sen, G.~Huerta, Y.~Deng, and K.~P. Bowman (2008), Error
  reduction and convergence in climate prediction, \textit{J. Climate},
  \textit{21}, 6698--6709, \doi{10.1175/2008JCLI2112.1}.

\bibitem[{\textit{Jakob}(2003)}]{Jakob03a}
Jakob, C. (2003), An improved strategy for the evaluation of cloud
  parameterizations in {GCMs}, \textit{Bull. Amer. Meteor. Soc.}, \textit{84},
  1387--1401, \doi{10.1175/BAMS-84-10-1387}.

\bibitem[{\textit{Jakob}(2010)}]{Jakob10a}
Jakob, C. (2010), Accelerating progress in global atmospheric model development
  through improved parameterizations: Challenges, opportunities, and
  strategies, \textit{Bull. Amer. Meteor. Soc.}, \textit{91}, 869--875,
  \doi{10.1175/2009BAMS2898.1}.

\bibitem[{\textit{J{\"a}rvinen et~al.}(2010)\textit{J{\"a}rvinen,
  R{\"a}is{\"a}nen, Laine, Tamminen, Ilin, Oja, Solonen, and
  Haario}}]{Jarvinen10a}
J{\"a}rvinen, H., P.~R{\"a}is{\"a}nen, M.~Laine, J.~Tamminen, A.~Ilin, E.~Oja,
  A.~Solonen, and H.~Haario (2010), Estimation of {ECHAM5} climate model
  closure parameters with adaptive {MCMC}, \textit{Atmos. Chem. Phys.},
  \textit{10}, 9993--10,002, \doi{10.5194/acp-10-9993-2010}.

\bibitem[{\textit{Jiang et~al.}(2012)\textit{Jiang, Su, Zhai, Perun, Genio,
  Nazarenko, Donner, Horowitz, Seman, Cole, Gettelman, Ringer, Rotstayn,
  Jeffrey, Wu, Brient, Dufresne, Kawai, Koshiro, Watanabe, L{\'E}cuyer,
  Volodin, Iversen, Drange, Mesquita, Read, Waters, Tian, Teixeira, and
  Stephens}}]{Jiang12a}
Jiang, J.~H., H.~Su, C.~Zhai, V.~S. Perun, A.~D. Genio, L.~S. Nazarenko, L.~J.
  Donner, L.~Horowitz, C.~Seman, J.~Cole, A.~Gettelman, M.~A. Ringer,
  L.~Rotstayn, S.~Jeffrey, T.~Wu, F.~Brient, J.-L. Dufresne, H.~Kawai,
  T.~Koshiro, M.~Watanabe, T.~S. L{\'E}cuyer, E.~M. Volodin, T.~Iversen,
  H.~Drange, M.~D.~S. Mesquita, W.~G. Read, J.~W. Waters, B.~Tian, J.~Teixeira,
  and G.~L. Stephens (2012), Evaluation of cloud and water vapor simulations in
  {CMIP5} climate models using {NASA} {``A-Train''} satellite observations,
  \textit{J. Geophys. Res.}, \textit{117}, D14,105, \doi{10.1029/2011JD017237}.

\bibitem[{\textit{Joiner et~al.}(2011)\textit{Joiner, Yoshida, Vasilkov, Corp,
  and Middleton}}]{Joiner11a}
Joiner, J., Y.~Yoshida, A.~Vasilkov, L.~Corp, and E.~Middleton (2011), First
  observations of global and seasonal terrestrial chlorophyll fluorescence from
  space, \textit{Biogeosci.}, \textit{8}, 637--651,
  \doi{10.5194/bg-8-637-2011}.

\bibitem[{\textit{Kaipio and Somersalo}(2005)}]{Kaipio05a}
Kaipio, J., and E.~Somersalo (2005), \textit{Statistical and Computational
  Inverse Problems}, vol. 160, Springer-Verlag, New York, NY.

\bibitem[{\textit{Karlsson and Svensson}(2013)}]{Karlsson13a}
Karlsson, J., and G.~Svensson (2013), Consequences of poor representation of
  {A}rctic sea-ice albedo and cloud-radiation interactions in the {CMIP5} model
  ensemble, \textit{Geophys. Res. Lett.}, \textit{40}, 4374--4379,
  \doi{10.1002/grl.50768}.

\bibitem[{\textit{Karlsson et~al.}(2008)\textit{Karlsson, Svensson, and
  Rodhe}}]{Karlsson08a}
Karlsson, J., G.~Svensson, and H.~Rodhe (2008), Cloud radiative forcing of
  subtropical low level clouds in global models, \textit{Clim. Dyn.},
  \textit{30}, 779--788, \doi{10.1007/s00382-007-0322-1}.

\bibitem[{\textit{Kasahara and Washington}(1967)}]{Kasahara67a}
Kasahara, A., and W.~M. Washington (1967), {NCAR} global general circulation
  model of the atmosphere, \textit{Mon. Wea. Rev.}, \textit{95}, 389--402.

\bibitem[{\textit{Kay et~al.}(2016)\textit{Kay, Wall, Yettella, Medeiros,
  Hannay, Caldwell, and Bitz}}]{Kay16a}
Kay, J.~E., C.~Wall, V.~Yettella, B.~Medeiros, C.~Hannay, P.~Caldwell, and
  C.~Bitz (2016), Global climate impacts of fixing the {S}outhern {O}cean
  shortwave radiation bias in the {C}ommunity {E}arth {S}ystem {M}odel
  {(CESM)}, \textit{J. Climate}, \textit{29}, 4617--4636,
  \doi{10.1175/JCLI-D-15-0358.1}.

\bibitem[{\textit{Kennedy and O'Hagan}(2001)}]{Kennedy01a}
Kennedy, M.~C., and A.~O'Hagan (2001), Bayesian calibration of computer models,
  \textit{J. Roy. Statist. Soc. B}, \textit{63}, 425--464,
  \doi{10.1111/1467-9868.00294}.

\bibitem[{\textit{Keppel-Aleks et~al.}(2012)\textit{Keppel-Aleks, Wennberg,
  Washenfelder, Wunch, Schneider, Toon, Andres, Blavier, Connor, Davis, Desai,
  Messerschmidt, Notholt, Roehl, Sherlock, Stephens, Vay, and
  Wofsy}}]{Keppel-Aleks12a}
Keppel-Aleks, G., P.~O. Wennberg, R.~A. Washenfelder, D.~Wunch, T.~Schneider,
  G.~C. Toon, R.~J. Andres, J.-F. Blavier, B.~Connor, K.~J. Davis, A.~R. Desai,
  J.~Messerschmidt, J.~Notholt, C.~M. Roehl, V.~Sherlock, B.~B. Stephens, S.~A.
  Vay, and S.~C. Wofsy (2012), The imprint of surface fluxes and transport on
  variations in total column carbon dioxide, \textit{Biogeosci.}, \textit{9},
  875--891.

\bibitem[{\textit{Khairoutdinov et~al.}(2005)\textit{Khairoutdinov, Randall,
  and DeMott}}]{Khairoutdinov05a}
Khairoutdinov, M., D.~Randall, and C.~DeMott (2005), Simulations of the
  atmospheric general circulation using a cloud-resolving model as a
  superparameterization of physical processes, \textit{J. Atmos. Sci.},
  \textit{62}, 2136--2154, \doi{10.1175/JAS3453.1}.

\bibitem[{\textit{Khairoutdinov and Randall}(2001)}]{Khairoutdinov01a}
Khairoutdinov, M.~F., and D.~A. Randall (2001), A cloud resolving model as a
  cloud parameterization in the {NCAR} {C}ommunity {C}limate {S}ystem {M}odel:
  {P}reliminary results, \textit{Geophys. Res. Lett.}, \textit{28}, 3617--3620.

\bibitem[{\textit{Khairoutdinov et~al.}(2009)\textit{Khairoutdinov, Krueger,
  Moeng, Bogenschutz, and Randall}}]{Khairoutdinov09a}
Khairoutdinov, M.~F., S.~K. Krueger, C.-H. Moeng, P.~A. Bogenschutz, and D.~A.
  Randall (2009), Large-eddy simulation of maritime deep tropical convection,
  \textit{J. Adv. Model. Earth Sys.}, \textit{1}, Art. \#15, 13 pp.,
  \doi{10.3894/JAMES.2009.1.15}.

\bibitem[{\textit{Klein and Majda}(2006)}]{Klein06a}
Klein, R., and A.~J. Majda (2006), Systematic multiscale models for deep
  convection on mesoscales, \textit{Theor. Comput. Fluid Dyn.}, \textit{20},
  525--551, \doi{10.1007/s00162-006-0027-9}.

\bibitem[{\textit{Klein and Hall}(2015)}]{Klein15a}
Klein, S.~A., and A.~Hall (2015), Emergent constraints for cloud feedbacks,
  \textit{Curr. Clim. Change Rep.}, \textit{1}, 276--287,
  \doi{10.1007/s40641-015-0027-1}.

\bibitem[{\textit{Klocke and Rodwell}(2014)}]{Klocke14a}
Klocke, D., and M.~J. Rodwell (2014), A comparison of two numerical weather
  prediction methods for diagnosing fast-physics errors in climate models,
  \textit{Quart. J. Roy. Meteor. Soc.}, \textit{140}, 517--524,
  \doi{10.1002/qj.2172}.

\bibitem[{\textit{Knorr}(2009)}]{Knorr09a}
Knorr, W. (2009), Is the airborne fraction of anthropogenic
  {CO\textsubscript{2}} emissions increasing?, \textit{Geophys. Res. Lett.},
  \textit{36}, L21,710, \doi{10.1029/2009GL040613}.

\bibitem[{\textit{Knutti et~al.}(2008)\textit{Knutti, Allen, Friedlingstein,
  Gregory, Hegerl, Meehl, Meinshausen, Murphy, Plattner, Raper, Stocker, Stott,
  Teng, and Wigley}}]{Knutti08b}
Knutti, R., M.~R. Allen, P.~Friedlingstein, J.~M. Gregory, G.~C. Hegerl, G.~A.
  Meehl, M.~Meinshausen, J.~M. Murphy, G.-K. Plattner, S.~C.~B. Raper, T.~F.
  Stocker, P.~A. Stott, H.~Teng, and T.~M.~L. Wigley (2008), A review of
  uncertainties in global temperature projections over the twenty-first
  century, \textit{J. Climate}, \textit{21}, 2651--2663,
  \doi{10.1175/2007JCLI2119.1}.

\bibitem[{\textit{K{\"o}hler et~al.}(2011)\textit{K{\"o}hler, Ahlgrimm, and
  Beljaars}}]{Kohler11a}
K{\"o}hler, M., M.~Ahlgrimm, and A.~Beljaars (2011), Unified treatment of dry
  convective and stratocumulus-topped boundary layers in the {ECMWF} model,
  \textit{Quart. J. Roy. Meteor. Soc.}, \textit{137}, 43--57,
  \doi{10.1002/qj.713}.

\bibitem[{\textit{Krasnopolsky et~al.}(2013)\textit{Krasnopolsky,
  Fox-Rabinovitz, and Belochitski}}]{Krasnopolsky13a}
Krasnopolsky, V.~M., M.~S. Fox-Rabinovitz, and A.~A. Belochitski (2013), Using
  ensemble of neural networks to learn stochastic convection parameterizations
  for climate and numerical weather prediction models from data simulated by a
  cloud resolving model, \textit{Adv. Art. Neur. Sys.}, \textit{2013}, 485,913,
  \doi{10.1155/2013/485913}.

\bibitem[{\textit{Lappen and Randall}(2001{\natexlab{a}})}]{Lappen01a}
Lappen, C.-L., and D.~A. Randall (2001{\natexlab{a}}), Toward a unified
  parameterization of the boundary layer and moist convection. {P}art~{I}: A
  new type of mass-flux model, \textit{J. Atmos. Sci.}, \textit{58},
  2021--2036.

\bibitem[{\textit{Lappen and Randall}(2001{\natexlab{b}})}]{Lappen01b}
Lappen, C.-L., and D.~A. Randall (2001{\natexlab{b}}), Toward a unified
  parameterization of the boundary layer and moist convection. {P}art~{II}:
  Lateral mass exchanges and subplume-scale fluxes, \textit{J. Atmos. Sci.},
  \textit{58}, 2037--2051.

\bibitem[{\textit{Law et~al.}(2015)\textit{Law, Stuart, and
  Zygalakis}}]{Law15a}
Law, K., A.~Stuart, and K.~Zygalakis (2015), \textit{Data Assimilation: A
  Mathematical Introduction}, \textit{Texts in Applied Mathematics}, vol.~62,
  Springer.

\bibitem[{\textit{Law and Stuart}(2012)}]{Law12a}
Law, K. J.~H., and A.~M. Stuart (2012), Evaluating data assimilation
  algorithms, \textit{Mon. Wea. Rev.}, \textit{140}, 3757--3782,
  \doi{10.1175/MWR-D-11-00257.1}.

\bibitem[{\textit{Le~Qu{\'e}r{\'e} et~al.}(2013)\textit{Le~Qu{\'e}r{\'e},
  Andres, Boden, Conway, Houghton, House, Marland, Peters, Van~der Werf,
  Ahlstr{\"o}m et~al.}}]{Le-Quere13a}
Le~Qu{\'e}r{\'e}, C., R.~J. Andres, T.~Boden, T.~Conway, R.~A. Houghton, J.~I.
  House, G.~Marland, G.~P. Peters, G.~Van~der Werf, A.~Ahlstr{\"o}m, et~al.
  (2013), The global carbon budget 1959--2011, \textit{Earth Sys. Sci. Data},
  \textit{5}, 165--185, \doi{doi:10.5194/essd-5-165-2013}.

\bibitem[{\textit{L'Ecuyer et~al.}(2015)\textit{L'Ecuyer, Beaudoing, Rodell,
  Olson, Lin, Kato, Clayson, Wood, Sheffield, Adler et~al.}}]{LEcuyer15a}
L'Ecuyer, T.~S., H.~K. Beaudoing, M.~Rodell, W.~Olson, B.~Lin, S.~Kato, C.~A.
  Clayson, E.~Wood, J.~Sheffield, R.~Adler, et~al. (2015), The observed state
  of the energy budget in the early twenty-first century, \textit{J. Climate},
  \textit{28}, 8319--8346, \doi{10.1175/JCLI-D-14-00556.1}.

\bibitem[{\textit{Li and Xie}(2014)}]{Li14a}
Li, G., and S.-P. Xie (2014), Tropical biases in {CMIP5} multi-model ensemble:
  The excessive equatorial {P}acific cold tongue and double {ITCZ} problems,
  \textit{J. Climate}, \textit{27}, 1765--1780,
  \doi{10.1175/JCLI-D-13-00337.1}.

\bibitem[{\textit{Lin}(2007)}]{Lin07a}
Lin, J.-L. (2007), The double-{ITCZ} problem in {IPCC} {AR4} coupled {GCM}s:
  Ocean--atmosphere feedback analysis, \textit{J. Climate}, \textit{20},
  4497--4525, \doi{10.1175/JCLI4272.1}.

\bibitem[{\textit{Lin et~al.}(2014)\textit{Lin, Qian, and Shinoda}}]{Lin14b}
Lin, J.-L., T.~Qian, and T.~Shinoda (2014), Stratocumulus clouds in
  {S}outheastern {P}acific simulated by eight {CMIP5}--{CFMIP} global climate
  models, \textit{J. Climate}, \textit{27}, 3000--3022,
  \doi{10.1175/JCLI-D-13-00376.1}.

\bibitem[{\textit{Liu et~al.}(2001)\textit{Liu, Moncrieff, and
  Grabowski}}]{Liu01a}
Liu, C., M.~W. Moncrieff, and W.~W. Grabowski (2001), Hierarchical modelling of
  tropical convective systems using explicit and parametrized approaches,
  \textit{Quart. J. Roy. Meteor. Soc.}, \textit{127}, 493--515.

\bibitem[{\textit{Liu et~al.}(2017)\textit{Liu, Bowman, Schimel, Parazoo,
  Jiang, Lee, Bloom, Wunch, Frankenberg, Sun, O'Dell, Gurney, Menemenlis,
  Gierach, Crisp, and Eldering}}]{Liu17a}
Liu, J., K.~W. Bowman, D.~S. Schimel, N.~C. Parazoo, Z.~Jiang, M.~Lee, A.~A.
  Bloom, D.~Wunch, C.~Frankenberg, Y.~Sun, C.~W. O'Dell, K.~R. Gurney,
  D.~Menemenlis, M.~Gierach, D.~Crisp, and A.~Eldering (2017), Contrasting
  carbon cycle responses of the tropical continents to the 2015--2016
  {E}l~{N}i{\~n}o, \textit{Science}, \textit{358}, eaam5690,
  \doi{10.1126/science.aam5690}.

\bibitem[{\textit{Lorenz}(1996)}]{Lorenz96a}
Lorenz, E.~N. (1996), Predictability---a problem partly solved, in
  \textit{Proc. Seminar on Predictability}, vol.~1, pp. 1--18, ECMWF, Reading,
  Berkshire, UK, {R}eprinted in T. N. Palmer and R. Hagedorn, eds.,
  \emph{Predictability of Weather and Climate}, Cambridge UP (2006).

\bibitem[{\textit{Lorenz and Emanuel}(1998)}]{Lorenz98a}
Lorenz, E.~N., and K.~A. Emanuel (1998), Optimal sites for supplementary
  weather observations: Simulation with a small model, \textit{J. Atmos. Sci.},
  \textit{55}, 399--414, \doi{10.1175/1520-0469(1998)055<0399:OSFSWO>2.0.CO;2}.

\bibitem[{\textit{Lucarini et~al.}(2014)\textit{Lucarini, Blender, Herbert,
  Ragone, Pascale, and Wouters}}]{Lucarini14a}
Lucarini, V., R.~Blender, C.~Herbert, F.~Ragone, S.~Pascale, and J.~Wouters
  (2014), Mathematical and physical ideas for climate science, \textit{Rev.
  Geophys.}, \textit{52}, 809--859, \doi{10.1002/2013RG000446}.

\bibitem[{\textit{Ma et~al.}(2013)\textit{Ma, Xie, Boyle, Klein, and
  Zhang}}]{Ma13a}
Ma, H.-Y., S.~Xie, J.~S. Boyle, S.~A. Klein, and Y.~Zhang (2013), Metrics and
  diagnostics for precipitation-related processes in climate model short-range
  hindcasts, \textit{J. Climate}, \textit{26}, 1516--1534,
  \doi{10.1175/JCLI-D-12-00235.1}.

\bibitem[{\textit{Majda}(2012)}]{Majda12a}
Majda, A.~J. (2012), Challenges in climate science and contemporary applied
  mathematics, \textit{Comm. Pure Appl. Math.}, \textit{65}, 920--948,
  \doi{10.1002/cpa.21401}.

\bibitem[{\textit{Majda et~al.}(2003)\textit{Majda, Timofeyev, and
  Vanden-Eijnden}}]{Majda03a}
Majda, A.~J., I.~Timofeyev, and E.~Vanden-Eijnden (2003), Systematic strategies
  for stochastic mode reduction in climate, \textit{J. Atmos. Sci.},
  \textit{60}, 1705--1722.

\bibitem[{\textit{Majda et~al.}(2008)\textit{Majda, Franzke, and
  Khouider}}]{Majda08a}
Majda, A.~J., C.~Franzke, and B.~Khouider (2008), An applied mathematics
  perspective on stochastic modelling for climate, \textit{Phil. Trans. R. Soc.
  A}, \textit{366}, 2429--2455, \doi{10.1098/rsta.2008.0012}.

\bibitem[{\textit{Manabe et~al.}(1965)\textit{Manabe, Smagorinsky, and
  Strickler}}]{Manabe65}
Manabe, S., J.~Smagorinsky, and R.~F. Strickler (1965), Simulated climatology
  of a general circulation model with a hydrologic cycle, \textit{Mon. Wea.
  Rev.}, \textit{93}, 769--798.

\bibitem[{\textit{Matheou and Chung}(2014)}]{Matheou14a}
Matheou, G., and D.~Chung (2014), Large-eddy simulation of stratified
  turbulence. {Part~II}: Application of the stretched-vortex model to the
  atmospheric boundary layer, \textit{J. Atmos. Sci.}, \textit{71}, 4439--4460,
  \doi{10.1175/JAS-D-13-0306.1}.

\bibitem[{\textit{Mauritsen et~al.}(2012)\textit{Mauritsen, Stevens, Roeckner,
  Crueger, Esch, Giorgetta, Haak, Jungclaus, Klocke, Matei, Mikolajewicz, Notz,
  Pincus, Schmidt, and Tomassini}}]{Mauritsen12a}
Mauritsen, T., B.~Stevens, E.~Roeckner, T.~Crueger, M.~Esch, M.~Giorgetta,
  H.~Haak, J.~Jungclaus, D.~Klocke, D.~Matei, U.~Mikolajewicz, D.~Notz,
  R.~Pincus, H.~Schmidt, and L.~Tomassini (2012), Tuning the climate of a
  global model, \textit{J. Adv. Model. Earth Sys.}, \textit{4}, M00A01,
  \doi{10.1029/2012MS000154}.

\bibitem[{\textit{Meinshausen et~al.}(2009)\textit{Meinshausen, Meinshausen,
  Hare, Raper, Frieler, Knutti, Frame, and Allen}}]{Meinshausen09a}
Meinshausen, M., N.~Meinshausen, W.~Hare, S.~C.~B. Raper, K.~Frieler,
  R.~Knutti, D.~J. Frame, and M.~R. Allen (2009), Greenhouse-gas emission
  targets for limiting global warming to {$2^\circ$C}, \textit{Nature},
  \textit{458}, 1158--1162, \doi{10.1038/nature08017}.

\bibitem[{\textit{Mintz}(1965)}]{Mintz65a}
Mintz, Y. (1965), Very long-term global integration of the primitive equations
  of atmospheric motion, in \textit{WMO-IUGG Symposium on Research and
  Development Aspects of Long-Range Forecasting, Boulder, Colo., 1964}, pp.
  141--155, World Meteorological Organization, Geneva.

\bibitem[{\textit{Moeng et~al.}(2007)\textit{Moeng, Dudhia, Klemp, and
  Sullivan}}]{Moeng07a}
Moeng, C.-H., J.~Dudhia, J.~Klemp, and P.~Sullivan (2007), Examining two-way
  grid nesting for large eddy simulation of the {PBL} using the {WRF} {M}odel,
  \textit{Mon. Wea. Rev.}, \textit{135}, 2295--2311, \doi{10.1175/MWR3406.1}.

\bibitem[{\textit{Nam et~al.}(2012)\textit{Nam, Bony, Dufresne, and
  Chepfer}}]{Nam12a}
Nam, C., S.~Bony, J.-L. Dufresne, and H.~Chepfer (2012), The `too few, too
  bright' tropical low-cloud problem in {CMIP5} models, \textit{Geophys. Res.
  Lett.}, \textit{39}, L21,801, \doi{10.1029/2012GL053421}.

\bibitem[{\textit{Neelin et~al.}(2009)\textit{Neelin, Peters, and
  Hales}}]{Neelin09a}
Neelin, J.~D., O.~Peters, and K.~Hales (2009), The transition to strong
  convection, \textit{J. Atmos. Sci.}, \textit{66}, 2367--2384.

\bibitem[{\textit{Neelin et~al.}(2010)\textit{Neelin, Bracco, Luo,
  {McWilliams}, and Meyerson}}]{Neelin10a}
Neelin, J.~D., A.~Bracco, H.~Luo, J.~C. {McWilliams}, and J.~E. Meyerson
  (2010), Considerations for parameter optimization and sensitivity in climate
  models, \textit{Proc. Natl. Acad. Sci.}, \textit{107}, 21,349--21,354,
  \doi{10.1073/pnas.1015473107}.

\bibitem[{\textit{Neggers et~al.}(2012)\textit{Neggers, Siebesma, and
  Heus}}]{Neggers12a}
Neggers, R. A.~J., A.~P. Siebesma, and T.~Heus (2012), Continuous single-column
  model evaluation at a permanent meteorological supersite, \textit{Bull. Amer.
  Meteor. Soc.}, \textit{93}, 1389--1400, \doi{10.1175/BAMS-D-11-00162.1}.

\bibitem[{\textit{Nie and Kuang}(2012)}]{Nie12a}
Nie, J., and Z.~Kuang (2012), Temperature and moisture perturbations: a
  comparison of large-eddy simulations and a convective parameterization based
  on stochastically entraining parcels, \textit{J. Atmos. Sci.}, \textit{69},
  in press.

\bibitem[{\textit{Nocedal and Wright}(2006)}]{Nocedal06a}
Nocedal, J., and S.~J. Wright (2006), \textit{Numerical Optimization}, Springer
  Series in Operations Research, 2nd ed., Springer.

\bibitem[{\textit{Ohno et~al.}(2016)\textit{Ohno, Satoh, and Yamada}}]{Ohno16a}
Ohno, T., M.~Satoh, and Y.~Yamada (2016), Warm cores, eyewall slopes, and
  intensities of tropical cyclones simulated by a 7-km-mesh global
  nonhydrostatic model, \textit{J. Atmos. Sci.}, \textit{73}, 4289--4309,
  \doi{10.1175/JAS-D-15-0318.1}.

\bibitem[{\textit{Ott et~al.}(2004)\textit{Ott, Hunt, Szunyogh, Zimin,
  Kostelich, Corazza, Kalnay, Patil, and Yorke}}]{Ott04a}
Ott, E., B.~R. Hunt, I.~Szunyogh, A.~V. Zimin, E.~J. Kostelich, M.~Corazza,
  E.~Kalnay, D.~J. Patil, and J.~A. Yorke (2004), A local ensemble {K}alman
  filter for atmospheric data assimilation, \textit{Tellus}, \textit{56},
  415--428, \doi{10.1111/j.1600-0870.2004.00076.x}.

\bibitem[{\textit{Palmer}(2014)}]{Palmer14c}
Palmer, T. (2014), Build high-resolution global climate models,
  \textit{Nature}, \textit{515}, 338--339, \doi{10.1038/515338a}.

\bibitem[{\textit{Palmer and Williams}(2010)}]{Palmer10a}
Palmer, T., and P.~Williams (Eds.) (2010), \textit{Stochastic Physics and
  Climate Modelling}, 480 pp., Cambridge Univ. Press.

\bibitem[{\textit{Palmer et~al.}(1998)\textit{Palmer, Gelaro, Barkmeijer, and
  Buizza}}]{Palmer98a}
Palmer, T.~N., R.~Gelaro, J.~Barkmeijer, and R.~Buizza (1998), Singular
  vectors, metrics, and adaptive observations, \textit{J. Atmos. Sci.},
  \textit{55}, 633--653, \doi{10.1175/1520-0469(1998)055<0633:SVMAAO>2.0.CO;2}.

\bibitem[{\textit{Palmer et~al.}(2005)\textit{Palmer, Shutts, Hagedorn,
  Doblas-Reyes, Jung, and Leutbecher}}]{Palmer05a}
Palmer, T.~N., G.~J. Shutts, R.~Hagedorn, F.~J. Doblas-Reyes, T.~Jung, and
  M.~Leutbecher (2005), Representing model uncertainty in weather and climate
  prediction, \textit{Annu. Rev. Earth Planet. Sci.}, \textit{33}, 163--193,
  \doi{10.1146/annurev.earth.33.092203.122552}.

\bibitem[{\textit{Parish and Duraisamy}(2016)}]{Parish16a}
Parish, E.~J., and K.~Duraisamy (2016), A paradigm for data-driven predictive
  modeling using field inversion and machine learning, \textit{J. Comp. Phys.},
  \textit{305}, 758--774, \doi{10.1016/j.jcp.2015.11.012}.

\bibitem[{\textit{Parishani et~al.}(2017)\textit{Parishani, Pritchard,
  Bretherton, Wyant, and Khairoutdinov}}]{Parishani17a}
Parishani, H., M.~S. Pritchard, C.~S. Bretherton, M.~C. Wyant, and
  M.~Khairoutdinov (2017), Toward low-cloud-permitting cloud
  superparameterization with explicit boundary layer turbulence, \textit{J.
  Adv. Model. Earth Sys.}, \textit{9}, 1542--1571, \doi{10.1002/2017MS000968}.

\bibitem[{\textit{Park}(2014{\natexlab{a}})}]{Park14a}
Park, S. (2014{\natexlab{a}}), A unified convection scheme ({UNICON}).
  {Part~I}: {F}ormulation, \textit{J. Atmos. Sci.}, \textit{71}, 3902--3930.

\bibitem[{\textit{Park}(2014{\natexlab{b}})}]{Park14b}
Park, S. (2014{\natexlab{b}}), A unified convection scheme ({UNICON}).
  {Part~II}: Simulation, \textit{J. Atmos. Sci.}, \textit{71}, 3931--3973.

\bibitem[{\textit{Phillips et~al.}(2004)\textit{Phillips, Potter, Williamson,
  Cederwall, Boyle, Fiorino, Hnilo, Olson, Xie, and Yio}}]{Phillips04a}
Phillips, T.~J., G.~L. Potter, D.~L. Williamson, R.~T. Cederwall, J.~S. Boyle,
  M.~Fiorino, J.~J. Hnilo, J.~G. Olson, S.~Xie, and J.~J. Yio (2004),
  Evaluating parameterizations in general circulation models: Climate
  simulation meets weather prediction, \textit{Bull. Amer. Meteor. Soc.},
  \textit{85}, 1903--1915, \doi{10.1175/BAMS-85-12-1903}.

\bibitem[{\textit{Pressel et~al.}(2015)\textit{Pressel, Kaul, Schneider, Tan,
  and Mishra}}]{Pressel15a}
Pressel, K.~G., C.~M. Kaul, T.~Schneider, Z.~Tan, and S.~Mishra (2015),
  Large-eddy simulation in an anelastic framework with closed water and entropy
  balances, \textit{J. Adv. Model. Earth Sys.}, \textit{7}, 1425--1456,
  \doi{10.1002/2015MS000496}.

\bibitem[{\textit{Pressel et~al.}(2017)\textit{Pressel, Mishra, Schneider,
  Kaul, and Tan}}]{Pressel17a}
Pressel, K.~G., S.~Mishra, T.~Schneider, C.~M. Kaul, and Z.~Tan (2017),
  Numerics and subgrid-scale modeling in large eddy simulations of
  stratocumulus clouds, \textit{J. Adv. Model. Earth Sys.}, \textit{9},
  1342--1365, \doi{10.1002/2016MS000778}.

\bibitem[{\textit{Pritchard and Somerville}(2009{\natexlab{a}})}]{Pritchard09a}
Pritchard, M.~S., and R.~C.~J. Somerville (2009{\natexlab{a}}), Empirical
  orthogonal function analysis of the diurnal cycle of precipitation in a
  multi-scale climate model, \textit{Geophys. Res. Lett.}, \textit{36},
  L05,812, \doi{10.1029/2008GL036964}.

\bibitem[{\textit{Pritchard and Somerville}(2009{\natexlab{b}})}]{Pritchard09b}
Pritchard, M.~S., and R.~C.~J. Somerville (2009{\natexlab{b}}), Assessing the
  diurnal cycle of precipitation in a multi-scale climate model, \textit{J.
  Adv. Model. Earth Sys..}, \textit{1}, \doi{10.3894/JAMES.2009.1.12}.

\bibitem[{\textit{Qu et~al.}(2014)\textit{Qu, Hall, Klein, and
  Caldwell}}]{Qu14a}
Qu, X., A.~Hall, S.~A. Klein, and P.~M. Caldwell (2014), On the spread of
  changes in marine low cloud cover in climate model simulations of the 21st
  century, \textit{Climate Dyn.}, \textit{42}, 2603--2626,
  \doi{10.1007/s00382-013-1945-z}.

\bibitem[{\textit{Qu et~al.}(2015)\textit{Qu, Hall, Klein, and
  DeAngelis}}]{Qu15a}
Qu, X., A.~Hall, S.~A. Klein, and A.~M. DeAngelis (2015), Positive tropical
  marine low-cloud cover feedback inferred from cloud-controlling factors,
  \textit{Geophys. Res. Lett.}, \textit{42}, \doi{10.1002/2015GL065627}.

\bibitem[{\textit{Randall}(2013)}]{Randall13b}
Randall, D.~A. (2013), Beyond deadlock, \textit{Geophys. Res. Lett.},
  \textit{40}, 5970--5976, \doi{10.1002/2013GL057998}.

\bibitem[{\textit{Randall and Wielicki}(1997)}]{Randall97a}
Randall, D.~A., and B.~A. Wielicki (1997), Measurements, models, and hypotheses
  in the atmospheric sciences, \textit{Bull. Amer. Meteor. Soc.}, \textit{78},
  400--406.

\bibitem[{\textit{Randall et~al.}(2003)\textit{Randall, Khairoutdinov, Arakawa,
  and Grabowski}}]{Randall03a}
Randall, D.~A., M.~Khairoutdinov, A.~Arakawa, and W.~Grabowski (2003), Breaking
  the cloud parameterization deadlock, \textit{Bull. Amer. Meteor. Soc.}, pp.
  1547--1564, \doi{10.1175/BAMS-84-11-1547}.

\bibitem[{\textit{Rodwell and Palmer}(2007)}]{Rodwell07a}
Rodwell, M.~J., and T.~N. Palmer (2007), Using numerical weather prediction to
  assess climate models, \textit{Quart. J. Roy. Meteor. Soc.}, \textit{133},
  129--146, \doi{10.1002/qj.23}.

\bibitem[{\textit{Romps}(2016)}]{Romps16a}
Romps, D.~M. (2016), The {S}tochastic {P}arcel {M}odel: A deterministic
  parameterization of stochastically entraining convection, \textit{J. Adv.
  Model. Earth Sys.}, \textit{8}, 319--344, \doi{10.1002/ 2015MS000537}.

\bibitem[{\textit{Romps and Kuang}(2010)}]{Romps10a}
Romps, D.~M., and Z.~Kuang (2010), Nature versus nurture in shallow convection,
  \textit{J. Atmos. Sci.}, \textit{67}, 1655--1666.

\bibitem[{\textit{Ruiz and Pulido}(2015)}]{Ruiz15a}
Ruiz, J., and M.~Pulido (2015), Parameter estimation using ensemble-based data
  assimilation in the presence of model error, \textit{Mon. Wea. Rev.},
  \textit{143}, 1568--1582, \doi{10.1175/MWR-D-14-00017.1}.

\bibitem[{\textit{Ruiz et~al.}(2013)\textit{Ruiz, Pulido, and
  Miyoshi}}]{Ruiz13a}
Ruiz, J.~J., M.~Pulido, and T.~Miyoshi (2013), Estimating model parameters with
  ensemble-based data assimilation: A review, \textit{J. Meteor. Soc. Japan},
  \textit{91}, 79--99, \doi{10.2151/jmsj.2013-201}.

\bibitem[{\textit{Schalkwijk et~al.}(2015)\textit{Schalkwijk, Jonker, Siebesma,
  and {Van Meijgaard}}}]{Schalkwijk15a}
Schalkwijk, J., H.~J.~J. Jonker, A.~P. Siebesma, and E.~{Van Meijgaard} (2015),
  Weather forecasting using {GPU}-based large-eddy simulations, \textit{Bull.
  Amer. Meteor. Soc.}, \textit{96}, 715--723, \doi{10.1175/BAMS-D-14-00114.1}.

\bibitem[{\textit{Schirber et~al.}(2013)\textit{Schirber, Klocke, Pincus,
  Quaas, and Anderson}}]{Schirber13a}
Schirber, S., D.~Klocke, R.~Pincus, J.~Quaas, and J.~L. Anderson (2013),
  Parameter estimation using data assimilation in an atmospheric general
  circulation model: From a perfect toward the real world, \textit{J. Adv.
  Model. Earth Sys.}, \textit{5}, 58--70, \doi{10.1029/2012MS000167}.

\bibitem[{\textit{Schneider et~al.}(2017)\textit{Schneider, Teixeira,
  Bretherton, Brient, Pressel, Sch{\"a}r, and Siebesma}}]{Schneider17a}
Schneider, T., J.~Teixeira, C.~S. Bretherton, F.~Brient, K.~G. Pressel,
  C.~Sch{\"a}r, and A.~P. Siebesma (2017), Climate goals and computing the
  future of clouds, \textit{Nature Climate Change}, \textit{7}, 3--5,
  \doi{10.1038/nclimate3190}.

\bibitem[{\textit{Schulthess}(2015)}]{Schulthess15a}
Schulthess, T.~C. (2015), Programming revisited, \textit{Nature Phys.},
  \textit{11}, 369--373.

\bibitem[{\textit{Shepherd et~al.}(2012)\textit{Shepherd, Ivins, Geruo,
  Barletta, Bentley, Bettadpur, Briggs, Bromwich, Forsberg, Galin, Horwath,
  Jacobs, Joughin, King, Lenaerts, Li et~al.}}]{Shepherd12a}
Shepherd, A., E.~R. Ivins, A.~Geruo, V.~R. Barletta, M.~J. Bentley,
  S.~Bettadpur, K.~H. Briggs, D.~H. Bromwich, R.~Forsberg, N.~Galin,
  M.~Horwath, S.~Jacobs, I.~Joughin, M.~A. King, J.~T.~M. Lenaerts, J.~Li,
  et~al. (2012), A reconciled estimate of ice-sheet mass balance,
  \textit{Science}, \textit{338}, 1183--1189, \doi{10.1126/science.1228102}.

\bibitem[{\textit{Siebesma et~al.}(2003)\textit{Siebesma, Bretherton, Brown,
  Chlond, Cuxart, Duynkerke, Jiang, Khairoutdinov, Lewellen, Moeng, Sanchez,
  Stevens, and Stevens}}]{Siebesma03}
Siebesma, A.~P., C.~S. Bretherton, A.~Brown, A.~Chlond, J.~Cuxart, P.~G.
  Duynkerke, H.~Jiang, M.~Khairoutdinov, D.~Lewellen, C.~H. Moeng, E.~Sanchez,
  B.~Stevens, and D.~E. Stevens (2003), A large eddy simulation intercomparison
  study of shallow cumulus convection, \textit{J. Atmos. Sci.}, \textit{60},
  1201--1219.

\bibitem[{\textit{Siebesma et~al.}(2007)\textit{Siebesma, Soares, and
  Teixeira}}]{Siebesma07}
Siebesma, A.~P., P.~M.~M. Soares, and J.~Teixeira (2007), A combined
  eddy-diffusivity mass-flux approach for the convective boundary layer,
  \textit{J. Atmos. Sci.}, \textit{64}, 1230--1248, \doi{10.1175/JAS3888.1}.

\bibitem[{\textit{Siler et~al.}(2017)\textit{Siler, Po-Chedley, and
  Bretherton}}]{Siler17a}
Siler, N., S.~Po-Chedley, and C.~S. Bretherton (2017), Variability in modeled
  cloud feedback tied to differences in the climatological spatial pattern of
  clouds, \textit{Clim. Dyn.}, \doi{10.1007/s00382-017-3673-2}.

\bibitem[{\textit{Simmons et~al.}(2016)\textit{Simmons, Fellous, Ramaswamy,
  Trenberth, Asrar, Balmaseda, Burrows, Ciais, Drinkwater, Friedlingstein,
  Gobron, Guilyardi, Halpern, Heimann, Johannessen, Levelt, Lopez-Baeza,
  Penner, Scholes, and Shepherd}}]{Simmons16a}
Simmons, A., J.-L. Fellous, V.~Ramaswamy, K.~Trenberth, G.~Asrar, M.~Balmaseda,
  J.~P. Burrows, P.~Ciais, M.~Drinkwater, P.~Friedlingstein, N.~Gobron,
  E.~Guilyardi, D.~Halpern, M.~Heimann, J.~Johannessen, P.~F. Levelt,
  E.~Lopez-Baeza, J.~Penner, R.~Scholes, and T.~Shepherd (2016), Observation
  and integrated {E}arth-system science: A roadmap for 2016--2025, \textit{Adv.
  Space Res.}, \textit{57}, 2037--2103, \doi{10.1016/j.asr.2016.03.008}.

\bibitem[{\textit{Smagorinsky}(1963)}]{Smagorinsky63}
Smagorinsky, J. (1963), General circulation experiments with the primitive
  equations. {I.} {T}he basic experiment, \textit{Mon. Wea. Rev.}, \textit{91},
  99--164.

\bibitem[{\textit{Smagorinsky et~al.}(1965)\textit{Smagorinsky, Manabe, and
  Holloway}}]{Smagorinsky65}
Smagorinsky, J., S.~Manabe, and J.~L. Holloway, Jr. (1965), Numerical results
  from a nine-level general circulation model of the atmosphere, \textit{Mon.
  Wea. Rev.}, \textit{93}, 727--768.

\bibitem[{\textit{Soden and Held}(2006)}]{Soden06a}
Soden, B.~J., and I.~M. Held (2006), An assessment of climate feedbacks in
  coupled ocean-atmosphere models, \textit{J. Climate}, \textit{19},
  3354--3360, \doi{10.1175/JCLI3799.1}.

\bibitem[{\textit{Solonen et~al.}(2012)\textit{Solonen, Ollinaho, Laine,
  Haario, Tamminen, and J{\"a}rvinen}}]{Solonen12a}
Solonen, A., P.~Ollinaho, M.~Laine, H.~Haario, J.~Tamminen, and H.~J{\"a}rvinen
  (2012), Efficient {MCMC} for climate model parameter estimation: Parallel
  adaptive chains and early rejection, \textit{Bayesian Anal.}, \textit{7},
  715--736, \doi{10.1214/12-BA724}.

\bibitem[{\textit{Stainforth et~al.}(2005)\textit{Stainforth, Aina,
  Christensen, Collins, Faull, Frame, Kettleborough, Knight, Martin, Murphy,
  Piani, Sexton, Smith, Spicer, Thorpe, and Allen}}]{Stainforth05a}
Stainforth, D.~A., T.~Aina, C.~Christensen, M.~Collins, N.~Faull, D.~J. Frame,
  J.~A. Kettleborough, S.~Knight, A.~Martin, J.~M. Murphy, C.~Piani, D.~Sexton,
  L.~A. Smith, R.~A. Spicer, A.~J. Thorpe, and M.~R. Allen (2005), Uncertainty
  in predictions of the climate response to rising levels of greenhouse gases,
  \textit{Nature}, \textit{433}, 403--406.

\bibitem[{\textit{Stan et~al.}(2010)\textit{Stan, Khairoutdinov, DeMott,
  Krishnamurthy, Straus, Randall, Kinter, and Shukla}}]{Stan10a}
Stan, C., M.~Khairoutdinov, C.~A. DeMott, V.~Krishnamurthy, D.~M. Straus, D.~A.
  Randall, J.~L. Kinter, and J.~Shukla (2010), An ocean-atmosphere climate
  simulation with an embedded cloud resolving model, \textit{Geophys. Res.
  Lett.}, \textit{37}, L01,702, \doi{10.1029/2009GL040822}.

\bibitem[{\textit{Stensrud}(2007)}]{Stensrud07a}
Stensrud, D.~J. (2007), \textit{Parameterization Schemes: Keys to Understanding
  Numerical Weather Prediction Models}, 477 pp., Cambridge Univ. Press,
  Cambridge, UK.

\bibitem[{\textit{Stephens et~al.}(2017)\textit{Stephens, Winker, Pelon,
  Trepte, Vane, Yuhas, L'Ecuyer, and Lebsock}}]{Stephens17a}
Stephens, G., D.~Winker, J.~Pelon, C.~Trepte, D.~Vane, C.~Yuhas, T.~L'Ecuyer,
  and M.~Lebsock (2017), {CloudSat} and {CALIPSO} within the {A}-{T}rain: {T}en
  years of actively observing the earth system, \textit{Bull. Amer. Meteor.
  Soc.}, \textit{in press}, \doi{10.1175/BAMS-D-16-0324.1}.

\bibitem[{\textit{Stephens}(2005)}]{Stephens05}
Stephens, G.~L. (2005), Cloud feedbacks in the climate system: A critical
  review, \textit{J. Climate}, \textit{18}, 237--273,
  \doi{10.1175/JCLI-3243.1}.

\bibitem[{\textit{Stephens et~al.}(2002)\textit{Stephens, Vane, Boain, Mace,
  Sassen, Wang, Illingworth, O'Connor, Rossow, Durden et~al.}}]{Stephens02a}
Stephens, G.~L., D.~G. Vane, R.~J. Boain, G.~G. Mace, K.~Sassen, Z.~Wang, A.~J.
  Illingworth, E.~J. O'Connor, W.~B. Rossow, S.~L. Durden, et~al. (2002), The
  {CloudSat} mission and the {A}-train, \textit{Bull. Amer. Meteor. Soc.},
  \textit{83}, 1771--1790, \doi{10.1175/BAMS-83-12-1771}.

\bibitem[{\textit{Stevens et~al.}(2005)\textit{Stevens, Moeng, Ackerman,
  Bretherton, Chlond, {de Roode}, Edwards, Golaz, Jiang, Khairoutdinov
  et~al.}}]{Stevens05a}
Stevens, B., C.-H. Moeng, A.~S. Ackerman, C.~S. Bretherton, A.~Chlond, S.~{de
  Roode}, J.~Edwards, J.-C. Golaz, H.~Jiang, M.~Khairoutdinov, et~al. (2005),
  Evaluation of large-eddy simulations via observations of nocturnal marine
  stratocumulus, \textit{Mon. Wea. Rev.}, \textit{133}, 1443--1462,
  \doi{10.1175/MWR2930.1}.

\bibitem[{\textit{Stewart et~al.}(2014)\textit{Stewart, Dance, Nichols, Eyre,
  and Cameron}}]{Stewart14a}
Stewart, L., S.~L. Dance, N.~K. Nichols, J.~Eyre, and J.~Cameron (2014),
  Estimating interchannel observation-error correlations for {IASI} radiance
  data in the {M}et {O}ffice system, \textit{Quart. J. Roy. Meteor. Soc.},
  \textit{140}, 1236--1244, \doi{10.1002/qj.2211}.

\bibitem[{\textit{Sun et~al.}(2017)\textit{Sun, Frankenberg, Wood, Schimel,
  Jung, Guanter, Drewry, Verma, Porcar-Castell, Griffis, Gu, Magney,
  K{\"o}hler, Evans, and Yuen}}]{Sun17a}
Sun, Y., C.~Frankenberg, J.~D. Wood, D.~S. Schimel, M.~Jung, L.~Guanter, D.~T.
  Drewry, M.~Verma, A.~Porcar-Castell, T.~J. Griffis, L.~Gu, T.~S. Magney,
  P.~K{\"o}hler, B.~Evans, and K.~Yuen (2017), {OCO-2} advances photosynthesis
  observation from space via solar-induced chlorophyll fluorescence,
  \textit{Science}, \textit{358}, eaam5747, \doi{10.1126/science.aam5747}.

\bibitem[{\textit{Suselj et~al.}(2013)\textit{Suselj, Teixeira, and
  Chung}}]{Suselj13a}
Suselj, K., J.~Teixeira, and D.~Chung (2013), A unified model for moist
  convective boundary layers based on a stochastic eddy-diffusivity/mass-flux
  parameterization, \textit{J. Atmos. Sci.}, \textit{70}, 1929--1953.

\bibitem[{\textit{Suzuki et~al.}(2013)\textit{Suzuki, Golaz, and
  Stephens}}]{Suzuki13a}
Suzuki, K., J.-C. Golaz, and G.~L. Stephens (2013), Evaluating cloud tuning in
  a climate model with satellite observations, \textit{Geophys. Res. Lett.},
  \textit{40}, 4463--4468, \doi{10.1002/grl.50874}.

\bibitem[{\textit{Swanson and Pierrehumbert}(1997)}]{Swanson97b}
Swanson, K.~L., and R.~T. Pierrehumbert (1997), Lower-tropospheric heat
  transport in the {P}acific storm track, \textit{J. Atmos. Sci.}, \textit{54},
  1533--1543.

\bibitem[{\textit{Tett et~al.}(2013)\textit{Tett, Mineter, Cartis, Rowlands,
  and Liu}}]{Tett13a}
Tett, S. F.~B., M.~J. Mineter, C.~Cartis, D.~J. Rowlands, and P.~Liu (2013),
  Can top-of-atmosphere radiation measurements constrain climate predictions?
  {Part~I}: {T}uning, \textit{J. Climate}, \textit{26}, 9348--9366,
  \doi{10.1175/JCLI-D-12-00595.1}.

\bibitem[{\textit{Tian}(2015)}]{Tian15a}
Tian, B. (2015), Spread of model climate sensitivity linked to
  double-intertropical convergence zone bias, \textit{Geophys. Res. Lett.},
  \textit{42}, 4133--4141, \doi{10.1002/2015GL064119}.

\bibitem[{\textit{Todd-Brown et~al.}(2013)\textit{Todd-Brown, Randerson, Post,
  Hoffman, Tarnocai, Schuur, and Allison}}]{Todd-Brown13a}
Todd-Brown, K., J.~Randerson, W.~Post, F.~Hoffman, C.~Tarnocai, E.~Schuur, and
  S.~Allison (2013), Causes of variation in soil carbon simulations from
  {CMIP5} {E}arth system models and comparison with observations,
  \textit{Biogeosci.}, \textit{10}, 1717--1736, \doi{10.5194/bg-10-1717-2013}.

\bibitem[{\textit{Vaughan et~al.}(2013)\textit{Vaughan, Comiso, Allison,
  Carrasco, Kaser, Kwok, Mote, Murray, Paul, Ren, Rignot, Solomina, Steffen,
  and Zhang}}]{Vaughan13a}
Vaughan, D.~G., J.~C. Comiso, I.~Allison, J.~Carrasco, G.~Kaser, R.~Kwok,
  P.~Mote, T.~Murray, F.~Paul, J.~Ren, E.~Rignot, O.~Solomina, K.~Steffen, and
  T.~Zhang (2013), Observations: Cryosphere, in \textit{Climate Change 2013:
  The Physical Science Basis. Contribution of Working Group I to the Fifth
  Assessment Report of the Intergovernmental Panel on Climate Change}, edited
  by T.~F. Stocker, D.~Qin, G.-K. Plattner, M.~Tignor, S.~K. Allen,
  J.~Boschung, A.~Nauels, Y.~Xia, V.~Bex, and P.~M. Midgley, chap.~4, pp.
  317--382, Cambridge University Press, Cambridge, UK, and New York, NY, USA.

\bibitem[{\textit{Vial et~al.}(2013)\textit{Vial, Dufresne, and
  Bony}}]{Vial13a}
Vial, J., J.-L. Dufresne, and S.~Bony (2013), On the interpretation of
  inter-model spread in {CMIP5} climate sensitivity estimates, \textit{Clim.
  Dyn.}, \textit{41}, 3339--3362, \doi{10.1007/s00382-013-1725-9}.

\bibitem[{\textit{Wan et~al.}(2014)\textit{Wan, Rasch, Zhang, Qian, Yan, and
  Zhao}}]{Wan14a}
Wan, H., P.~J. Rasch, K.~Zhang, Y.~Qian, H.~Yan, and C.~Zhao (2014), Short
  ensembles: an efficient method for discerning climate-relevant sensitivities
  in atmospheric general circulation models, \textit{Geosci. Model Dev..},
  \textit{7}, 1961--1977, \doi{10.5194/gmd-7-1961-2014}.

\bibitem[{\textit{Wang et~al.}(2014)\textit{Wang, Hu, and Blonigan}}]{Wang14c}
Wang, Q., R.~Hu, and P.~Blonigan (2014), Least {S}quares {S}hadowing
  sensitivity analysis of chaotic limit cycle oscillations, \textit{J. Comp.
  Phys.}, \textit{267}, 210--224, \doi{10.1016/j.jcp.2014.03.002}.

\bibitem[{\textit{Webb et~al.}(2001)\textit{Webb, Senior, Bony, and
  Morcrette}}]{Webb01a}
Webb, M., C.~Senior, S.~Bony, and J.-J. Morcrette (2001), Combining {ERBE} and
  {ISCCP} data to assess clouds in the {H}adley {C}entre, {ECMWF} and {LMD}
  atmospheric climate models atmospheric climate models, \textit{Clim. Dyn.},
  \textit{17}, 905--922.

\bibitem[{\textit{Webb et~al.}(2013)\textit{Webb, Lambert, and
  Gregory}}]{Webb13b}
Webb, M.~J., F.~H. Lambert, and J.~M. Gregory (2013), Origins of differences in
  climate sensitivity, forcing and feedback in climate models, \textit{Clim.
  Dyn.}, \textit{40}, 677--707, \doi{10.1007/s00382-012-1336-x}.

\bibitem[{\textit{Wenzel et~al.}(2014)\textit{Wenzel, Cox, Eyring, and
  Friedlingstein}}]{Wenzel14a}
Wenzel, S., P.~M. Cox, V.~Eyring, and P.~Friedlingstein (2014), Emergent
  constraints on climate-carbon cycle feedbacks in the {CMIP5} {E}arth system
  models, \textit{Biogeosci.}, \textit{119}, 794--807.

\bibitem[{\textit{Wenzel et~al.}(2016)\textit{Wenzel, Cox, Eyring, and
  Friedlingstein}}]{Wenzel16a}
Wenzel, S., P.~M. Cox, V.~Eyring, and P.~Friedlingstein (2016), Projected land
  photosynthesis constrained by changes in the seasonal cycle of atmospheric
  {CO}\textsubscript{2}, \textit{Nature}, \textit{538}, 499--501,
  \doi{10.1038/nature19772}.

\bibitem[{\textit{Wilks}(2005)}]{Wilks05a}
Wilks, D.~S. (2005), Effects of stochastic parametrizations in the {L}orenz '96
  system, \textit{Quart. J. Roy. Meteor. Soc.}, \textit{131}, 389--407,
  \doi{10.1256/qj.04.03}.

\bibitem[{\textit{Wood}(2012)}]{Wood12a}
Wood, R. (2012), Stratocumulus clouds, \textit{Mon. Wea. Rev.}, \textit{140},
  2373--2423, \doi{10.1175/MWR-D-11-00121.1}.

\bibitem[{\textit{Wouters and Lucarini}(2013)}]{Wouters13a}
Wouters, J., and V.~Lucarini (2013), Multi-level dynamical systems: Connecting
  the {R}uelle response theory and the {M}ori-{Z}wanzig approach, \textit{J.
  Stat. Phys.}, \textit{151}, 850--860, \doi{10.1007/s10955-013-0726-8}.

\bibitem[{\textit{Wouters et~al.}(2016)\textit{Wouters, Dolaptchiev, and
  Lucarini}}]{Wouters16b}
Wouters, J., S.~I. Dolaptchiev, and V.~Lucarini (2016), Parameterization of
  stochastic multiscale triads, \textit{Nonlin. Processes Geophys.},
  \textit{23}, 435--445, \doi{10.5194/npg-23-435-2016}.

\bibitem[{\textit{Xie et~al.}(2012)\textit{Xie, Ma, Boyle, Klein, and
  Zhang}}]{Xie12a}
Xie, S., H.-Y. Ma, J.~S. Boyle, S.~A. Klein, and Y.~Zhang (2012), On the
  correspondence between short- and long-time-scale systematic errors in
  {CAM4}/{CAM5} for the {Y}ear of {T}ropical {C}onvection, \textit{J. Climate},
  \textit{25}, 7937--7955, \doi{10.1175/JCLI-D-12-00134.1}.

\bibitem[{\textit{Yokota et~al.}(2009)\textit{Yokota, Yoshida, Eguchi, Ota,
  Tanaka, Watanabe, and Maksyutov}}]{Yokota09a}
Yokota, T., Y.~Yoshida, N.~Eguchi, Y.~Ota, T.~Tanaka, H.~Watanabe, and
  S.~Maksyutov (2009), Global concentrations of {CO\textsubscript{2}} and
  {CH\textsubscript{4}} retrieved from {GOSAT}: First preliminary results,
  \textit{SOLA}, \textit{5}, 160--163.

\bibitem[{\textit{Zhang et~al.}(2005)\textit{Zhang, Lin, Klein, Bacmeister,
  Bony, Cederwall, Del~Genio, Hack, Loeb, Lohmann, Minnis, Musat, Pincus
  et~al.}}]{Zhang05b}
Zhang, M.~H., W.~Y. Lin, S.~A. Klein, J.~T. Bacmeister, S.~Bony, R.~T.
  Cederwall, A.~D. Del~Genio, J.~J. Hack, N.~G. Loeb, U.~Lohmann, P.~Minnis,
  I.~Musat, R.~Pincus, et~al. (2005), Comparing clouds and their seasonal
  variations in 10 atmospheric general circulation models with satellite
  measurements, \textit{J. Geophys. Res.}, \textit{110}, D15S02,
  \doi{10.1029/2004JD005021}.

\bibitem[{\textit{Zhang et~al.}(2015)\textit{Zhang, Liu, and Zhang}}]{Zhang15a}
Zhang, X., H.~Liu, and M.~Zhang (2015), Double {ITCZ} in coupled
  ocean-atmosphere models: From {CMIP3} to {CMIP5}, \textit{Geophys. Res.
  Lett.}, \textit{42}, 8651--8659, \doi{10.1002/2015GL065973}.

\bibitem[{\textit{Zhao et~al.}(2016)\textit{Zhao, Held, Ramaswamy, Lin, Ming,
  Ginoux, Wyman, Donner, and Paynter}}]{Zhao16a}
Zhao, M., I.~M. Held, V.~Ramaswamy, S.-J. Lin, Y.~Ming, P.~Ginoux, B.~Wyman,
  L.~J. Donner, and D.~Paynter (2016), Uncertainty in model climate sensitivity
  traced to representations of cumulus precipitation microphysics, \textit{J.
  Climate}, \textit{29}, 543--560, \doi{10.1175/JCLI-D-15-0191.1}.

\bibitem[{\textit{Zhu et~al.}(2010)\textit{Zhu, Albrecht, Ghate, and
  Zhu}}]{Zhu10a}
Zhu, P., B.~A. Albrecht, V.~P. Ghate, and Z.~Zhu (2010), Multiple-scale
  simulations of stratocumulus clouds, \textit{J. Geophys. Res.}, \textit{115},
  D23,201, \doi{10.1029/2010JD014400}.

\end{thebibliography}
\end{document}